%% file: main.tex
\documentclass{article} 
\usepackage{iclr2024_conference,times}

\input{math_commands.tex}

\usepackage[
colorlinks=true,
breaklinks=true,
urlcolor=magenta,
linkcolor=red,
citecolor=blue,
]{hyperref}
\usepackage{url}

\usepackage{booktabs}     
\usepackage{amsfonts}     
\usepackage{nicefrac}     
\usepackage{microtype}    
\usepackage{xcolor}       
\usepackage{comment}  
\usepackage{graphicx}
\usepackage{subcaption}
\usepackage{natbib}
\usepackage{xcolor}
\usepackage{threeparttable}
\usepackage{enumitem}
\usepackage{adjustbox}
\usepackage{tabularray}
\usepackage{nicematrix}
\usepackage{pifont}
\newcommand{\cmark}{\ding{51}}
\newcommand{\xmark}{\ding{55}}
\usepackage{multirow}
\usepackage{wrapfig}

\usepackage{tablefootnote}

\usepackage{titlesec} 
\titlespacing*{\paragraph} {0pt}{2pt}{4pt}

\title{Soft Contrastive Learning for Time Series}
\author{Seunghan Lee, Taeyoung Park, Kibok Lee 
\\
Department of Statistics and Data Science, Yonsei University\\  
\texttt{\{seunghan9613,tpark,kibok\}@yonsei.ac.kr} \\
}

\iclrfinalcopy 
\begin{document}

\maketitle

\begin{abstract}
Contrastive learning has shown to be effective to learn representations from time series in a self-supervised way.
However, contrasting similar time series instances or values from adjacent timestamps within a time series leads to ignore their inherent correlations, which results in deteriorating the quality of learned representations.
To address this issue, we propose \textit{SoftCLT}, a simple yet effective soft contrastive learning strategy for time series.
This is achieved by introducing instance-wise and temporal contrastive loss with soft assignments ranging from zero to one.
Specifically, we define soft assignments for 1) instance-wise contrastive loss by the distance between time series on the data space, and 2) temporal contrastive loss by the difference of timestamps.
SoftCLT is a plug-and-play method for time series contrastive learning that improves the quality of learned representations without bells and whistles.
In experiments, we demonstrate that SoftCLT consistently improves the performance in various downstream tasks including classification, semi-supervised learning, transfer learning, and anomaly detection, showing state-of-the-art performance.
Code is available at this repository: \url{https://github.com/seunghan96/softclt}.
\end{abstract}

\section{Introduction}
Time series (TS) data are 
ubiquitous in many fields, including finance, energy, healthcare, and transportation~\citep{ding2020hierarchical, lago2018forecasting, solares2020deep, cai2020traffic}.
However, annotating TS data can be challenging as it often requires significant domain expertise and time.
To overcome the limitation and utilize unlabeled data without annotations, self-supervised learning has emerged as a promising
representation learning
approach not only in 
natural language processing~\citep{devlin2018bert, gao2021simcse} and 
computer vision~\citep{chen2020simple, dosovitskiy2020image}, 
but also in TS analysis~\citep{franceschi2019unsupervised, yue2022ts2vec}.
In particular, contrastive learning (CL) has demonstrated remarkable performance across different domains~\citep{chen2020simple, gao2021simcse, yue2022ts2vec}.
As it is challenging to determine similarities of instances in self-supervised learning,
recent CL works apply data augmentation to generate two views per data and take views from the same instance as positive pairs and the others as negatives~\citep{chen2020simple}.
However, we argue that the standard CL objective might be harmful for TS representation learning,
because inherent correlations in similar TS instances and values nearby timestamps within a TS, which could be a strong self-supervision, are ignored in CL.
For example, 
distance metrics such as dynamic time warping (DTW) have been widely used for 
measuring the similarities of TS data,
and contrasting TS data might lose such information.
Also, values with close timestamps are usually similar in natural TS data, so contrasting all values with different timestamps with the same degree of penalty as in previous CL methods~\citep{eldele2021time, yue2022ts2vec} might not be optimal.
Motivated by this, we explore the following research question:
\textit{how can we take account of the similarities of time series data for better contrastive representation learning?}
To this end, we propose \textbf{Soft} \textbf{C}ontrastive \textbf{L}earning for \textbf{T}ime series (\textit{SoftCLT}).
Specifically, we propose to consider the InfoNCE loss~\citep{oord2018representation} 
not only for the positive pairs but also all other pairs and compute their weighted summation in both instance-wise CL and temporal CL, where
instance-wise CL contrasts the representations of TS instances,
while temporal CL contrasts the representations of timestamps within a single TS, as shown in Figure~\ref{fig:pair}.
We propose to assign soft assignments based on the distance between TS for the instance-wise CL, and the difference of timestamps for the temporal CL.
This formulation can be seen as a generalization of the standard contrastive loss, 
as the proposed loss
becomes the contrastive loss if we replace soft assignments with hard assignments of either zero for negative or one for positive.

We conduct extensive experiments in various tasks, including TS classification, semi-supervised classification, transfer learning, and anomaly detection tasks to prove the effectiveness of the proposed method.
Experimental results validate that our method improves the performance of previous CL methods, achieving state-of-the-art (SOTA) performance on a range of downstream tasks. 
The main contributions of this paper are summarized as follows:
\setlist[itemize]{leftmargin=0.3cm}
\begin{itemize}
    \item We propose SoftCLT, a simple yet effective soft contrastive learning strategy for TS. 
    Specifically, we propose soft contrastive losses for instance and temporal dimensions, respectively, to address limitations of previous CL methods for TS.
    \item We provide extensive experimental results on various tasks for TS, showing that our method improves SOTA performance on a range of downstream tasks.
    For example, SoftCLT improves the average accuracy of 125 UCR datasets and 29 UEA datasets by 2.0\% and 3.9\%, respectively, compared to the SOTA unsupervised representation for classification tasks.
    \item SoftCLT is easily applicable to other CL frameworks for TS by introducing soft assignments
    and its overhead is negligible, making it practical for use.
\end{itemize}

\section{Related Work}
\textbf{Self-supervised learning. }
In recent years, self-supervised learning has gained lots of attention
for its ability to learn powerful representations from large amounts of unlabeled data. 
Self-supervised learning is done by training a model to solve a pretext task derived from a certain aspect of data without supervision.
As a self-supervised pretext task, next token prediction~\citep{brown2020language} and masked token prediction~\citep{devlin2018bert} are commonly used in natural language processing, while solving jigsaw puzzles~\citep{Noroozi2016Jigsaw} and rotation prediction~\citep{gidaris2018unsupervised} are proposed in computer vision.
In particular, contrastive learning~\citep{hadsell2006dimensionality} has shown to be an effective pretext task across domains, which maximizes similarities of positive pairs while minimizing similarities of negative pairs~\citep{gao2021simcse, chen2020simple, yue2022ts2vec}.
\begin{table}[t]
\vspace{-12pt}
\centering
\begin{adjustbox}{max width=0.999\columnwidth}
\begin{NiceTabular}{c|c|c|c|c|c|c|c|c|c|c|c|c} \toprule
 & T-Loss & Self-Time & TNC & TS-SD & TS-TCC & TS2Vec & Mixing-Up & CoST & TimeCLR & TF-C & CA-TCC & SoftCLT \\
& (NeurIPS \citeyear{franceschi2019unsupervised})& (arxiv \citeyear{fan2020self}) & (ICLR \citeyear{tonekaboni2021unsupervised}) & (IJCNN \citeyear{shi2021self}) & (IJCAI \citeyear{eldele2021time}) & (AAAI \citeyear{yue2022ts2vec}) & (PR Letters \citeyear{wickstrom2022mixing}) & (ICLR \citeyear{woo2022cost}) & (KBS \citeyear{yang2022timeclr}) & (NeurIPS \citeyear{zhang2022self}) & (TPAMI \citeyear{eldele2022self}) &  (Ours) \\
\midrule Instanse-wise CL & \cmark & \cmark &  & \cmark & \cmark & \cmark & \cmark & \cmark & \cmark & \cmark & \cmark & \cmark\\
\midrule Temporal CL &  & \cmark & \cmark &  & \cmark & \cmark & &  &  &  & \cmark & \cmark\\
\midrule Hierarchical CL &  &  &  & & & \cmark & & &  & &  & \cmark \\
\midrule Soft CL &  &  &  & & & & &  & & & & \cmark \\
\bottomrule
\end{NiceTabular}
\end{adjustbox}
\vspace{-6pt}
\caption{Comparison table of contrastive learning methods in time series.}
\label{tb;tscl-compare}
\vspace{-15pt}
\end{table} 

\textbf{Contrastive learning in time series. }
In the field of TS analysis, several designs for positive and negative pairs have been proposed for CL, taking into account the invariant properties of TS.
Table~\ref{tb;tscl-compare} compares various CL methods in TS including ours in terms of several properties.
T-Loss~\citep{franceschi2019unsupervised} samples a random subseries from a TS and treats them as positive when they belong to its subseries, and negative if belong to subseries of other TS.
Self-Time~\citep{fan2020self} captures inter-sample relation between TS by defining augmented sample of same TS as positive and negative otherwise, and captures intra-temporal relation within TS by solving a classification task, where the class labels are defined using the temporal distance between the subseries.
TNC~\citep{tonekaboni2021unsupervised} defines temporal neighborhood of windows using normal distribution and treats samples in neighborhood as positives.
TS-SD~\citep{shi2021self} trains a model using triplet similarity discrimination task, where the goal is to identify which of two TS is more similar to a given TS, using DTW to define similarity.
TS-TCC~\citep{eldele2021time} proposes a temporal contrastive loss by making the augmentations predict each other's future, and CA-TCC~\citep{eldele2022self}, which is the extension of TS-TCC to the semi-supervised setting, adopts the same loss.
TS2Vec~\citep{yue2022ts2vec} splits TS into two subseries and defines hierarchical contrastive loss in both instance and temporal dimensions.
Mixing-up~\citep{wickstrom2022mixing} generates new TS by mixing two TS, where the goal is to predict the mixing weights.
CoST~\citep{woo2022cost} utilizes both time domain and frequency domain contrastive losses to learn disentangled seasonal-trend representations of TS.
TimeCLR~\citep{yang2022timeclr} introduces phase-shift and amplitude change augmentations, which are data augmentation methods based on DTW.
TF-C~\citep{zhang2022self} learns both time- and frequency-based representations of TS and proposes a novel time-frequency consistency architecture. 
In the medical domain,
Subject-Aware CL \citep{cheng2020subject} proposes an instance-wise CL framework where the temporal information is entangled by architecture design, and
CLOCS \citep{kiyasseh2021clocs} proposes to consider spatial dimension specifically available in their application, which is close to the channels in general TS.
While previous CL methods for TS compute \textit{hard} contrastive loss, where the similarities between all negative pairs are equally minimized, we introduce \textit{soft} contrastive loss for TS.

\textbf{Soft contrastive learning. }
CL is typically done by batch instance discrimination, where each instance is considered to be in a distinct class.
However, this approach can pose a risk of pushing similar samples farther apart in the embedding space. 
To address this issue, several methods have been proposed, including a method that utilizes soft assignments of images~\citep{thoma2020soft} based on feature distances and geometric proximity measures.
NNCLR~\citep{dwibedi2021little} defines additional positives for each view by extracting top-$k$ neighbors in the feature space. 
NCL \citep{yeche2021neighborhood} finds neighbors
using supervision from the medical domain knowledge
and jointly optimizes two conflicting losses with a trade-off:
the neighbor alignment loss maximizing the similarity of neighbors as well as positive pairs, and
the neighbor discriminative loss maximizing the similarity of positive pairs while minimizing the similarity of neighbors.
SNCLR~\citep{ge2023soft}, which extends NNCLR with soft assignments,
employs an attention module to determine the correlations between the current and neighboring samples.
CO2~\citep{wei2020co2} introduces consistency regularization to enforce relative distribution consistency between different positive views and all negatives, resulting in soft relationships between samples. 
ASCL~\citep{feng2022adaptive} introduces soft inter-sample relations by transforming the original instance discrimination task into a multi-instance soft discrimination task.
Previous soft CL methods in non-TS domains compute soft assignments on the \textit{embedding space}, because similarities of instances on the data space are difficult to measure,
particularly in computer vision~\citep{chen2020simple}.
In contrast, we propose to compute soft assignments based on the distance between TS instances on the \textit{data space}.

\textbf{Masked modeling in time series. }
Other than CL, masked modeling has recently been studied as a pretext task for self-supervised learning in TS by masking out a portion of TS and predicting the missing values.
While CL has demonstrated remarkable performance in high-level classification tasks, masked modeling has excelled in low-level forecasting tasks~\citep{dong2023simmtm,huang2022contrastive,xie2022simmim}.
TST~\citep{zerveas2021transformer} adopts the masked modeling paradigm to TS, where the goal is to reconstruct the masked timestamps. 
PatchTST~\citep{nie2022time} aims to predict the masked subseries-level patches to capture the local semantic information and reduce memory usage. 
SimMTM~\citep{dong2023simmtm} reconstructs the original TS from multiple masked TS.

\section{Methodology}
\begin{figure*}[t]
\vspace{-12pt}
\centering
\includegraphics[width=0.99 \textwidth]{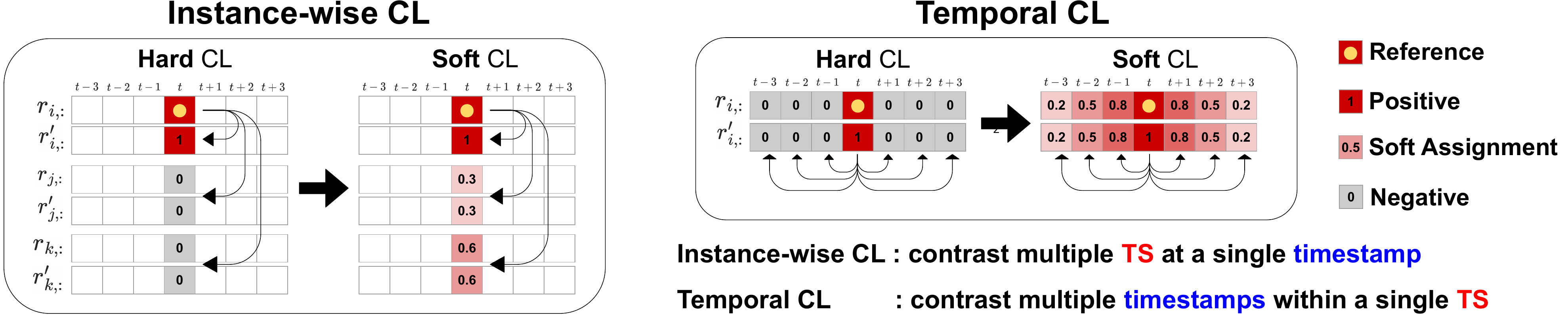} 
\caption{\textbf{Overall framework of SoftCLT. } Unlike the conventional hard CL that gives either positive or negative assignments to sample pairs, SoftCLT gives soft assignments to both instance-wise and temporal relationships. 
Two views of the same sample are denoted as $r$ and $\tilde{r}$, respectively.}
\label{fig:pair}
\vspace{-12pt}
\end{figure*}

In this section, we propose SoftCLT by introducing soft assignments to instance-wise and temporal contrastive losses to capture both inter-sample and intra-temporal relationships, respectively.
For instance-wise CL, we use distance between TS on the data space to capture the inter-sample relations, and for temporal CL, we use the difference between timestamps to consider the temporal relation within a single TS.
The overall framework of SoftCLT is illustrated in Figure~\ref{fig:pair}.

\subsection{Problem Definition}
This paper addresses the task of learning a nonlinear embedding function $f_{\theta}: x \rightarrow r$, given a batch of $N$ time series $\mathcal{X}=\left\{x_1, \hdots, x_N\right\}$.
Our goal is to learn $f_{\theta}$ mapping a time series $x_i \in \mathbb{R}^{T \times D}$ to a representation vector $r_i = \left[ r_{i, 1}, \hdots, r_{i, T} \right]^\top \in \mathbb{R}^{T \times M}$, where $T$ is the sequence length, $D$ is the input feature dimension, and $M$ is the embedded feature dimension.

\subsection{Soft Instance-Wise Contrastive Learning}
Contrasting all instances within a batch might be harmful for TS representation learning because similar instances are learned to be far away from each other on the embedding space.
Unlike other domains such as computer vision,
the distance between TS data computed on the data space is useful for measuring the similarity of them.
For example, the pixel-by-pixel distance of two different images is not related to their similarities in general, that of two TS data is useful to measure their similarities.
With a min-max normalized distance metric $D(\cdot, \cdot)$,
we define a soft assignment for a pair of data indices $(i,i^{\prime})$ for the instance-wise contrastive loss
using the sigmoid function $\sigma(a) = 1/(1+\exp(-a))$:
\begin{equation}{
    w_{I}(i,i^{\prime}) = 2 \alpha \cdot \sigma \left(  - \tau_{I} \cdot D(x_i,x_{i^{\prime}}) \right),
    \label{eqn:inst_weight}
}
\end{equation}
where $\tau_{I}$ is a hyperparameter controlling the sharpness
and $\alpha$ is the upper bound in the range of $[0,1]$
to distinguish pairs of the same TS and pairs of different TS close to each other;
when $\alpha=1$, we give the assignment of one to the pairs with the distance of zero as well as the pairs of the same TS.
Note that distances between TS are computed with the \textit{original} TS rather than the augmented views, because the pairwise distance matrix can be precomputed offline or cached for efficiency.

For the choice of the distance metric $D$, we conduct an ablation study in Table~\ref{tbl:ablation4}, comparing 1) cosine distance, 2) Euclidean distance, 3) dynamic time warping (DTW), and 4) time alignment measurement (TAM)~\citep{folgado2018time}.
Among them, we choose DTW as the distance metric throughout the experiments based on the result in Table~\ref{tbl:ablation4}.
While the computational complexity of DTW is $\mathcal{O}(T^2)$ for two TS of length $T$ which might be costly for large-scale datasets, it can be precomputed offline or cached to facilitate efficient calculations, or its fast version such as FastDTW~\citep{salvador2007toward} with the complexity of $\mathcal{O}(T)$ can be used.
We empirically confirmed that the output of DTW and FastDTW is almost the same, such that the CL results also match.

Let $r_{i,t}=r_{i+2N,t}$ and $\tilde{r}_{i,t} = r_{i+N,t}$ be the embedding vectors from two augmentations of $x_{i}$ at timestamp $t$ for conciseness.
Inspired by the fact that the contrastive loss can be interpreted as the cross-entropy loss~\citep{lee2021mix}, we define a softmax probability of the relative similarity out of all similarities considered when computing the loss as:
\begin{equation}
    p_{I}((i,i^{\prime}),t) = \frac
    {\exp (r_{i,t} \circ r_{i^{\prime},t})}
    {\sum_{j=1, j\neq i}^{2N} \exp (r_{i, t} \circ r_{j, t})},
    \label{eqn:inst_prob}
\end{equation}
where we use the dot product as the similarity measure $\circ$.
Then, the soft instance-wise contrastive loss for
$x_{i}$ at timestamp $t$ is defined as:
\begin{equation}
{
    \ell_{I}^{(i, t)} = 
    - \log p_{I}((i,i+N),t)
    - \sum_{j=1,j \neq \{i, i+N\}}^{2N} w_{I}(i,j \text{ mod } N) \cdot \log p_{I}((i,j),t).
    \label{eqn:inst_loss}
}
\end{equation}
The first term in $\ell_{I}^{(i, t)}$ corresponds to the loss of the positive pair, and the second term corresponds to that of the other pairs weighted by soft assignments $w_{I}(i,i^{\prime})$.
Note that this loss can be seen as a generalization of the hard instance-wise contrastive loss, which is the case when $\forall w_{I}(i,i^{\prime})=0$.

\subsection{Soft Temporal Contrastive Learning}
\begin{figure}[t]
  \centering
  \begin{subfigure}{0.545\textwidth}
    \includegraphics[width=\linewidth]{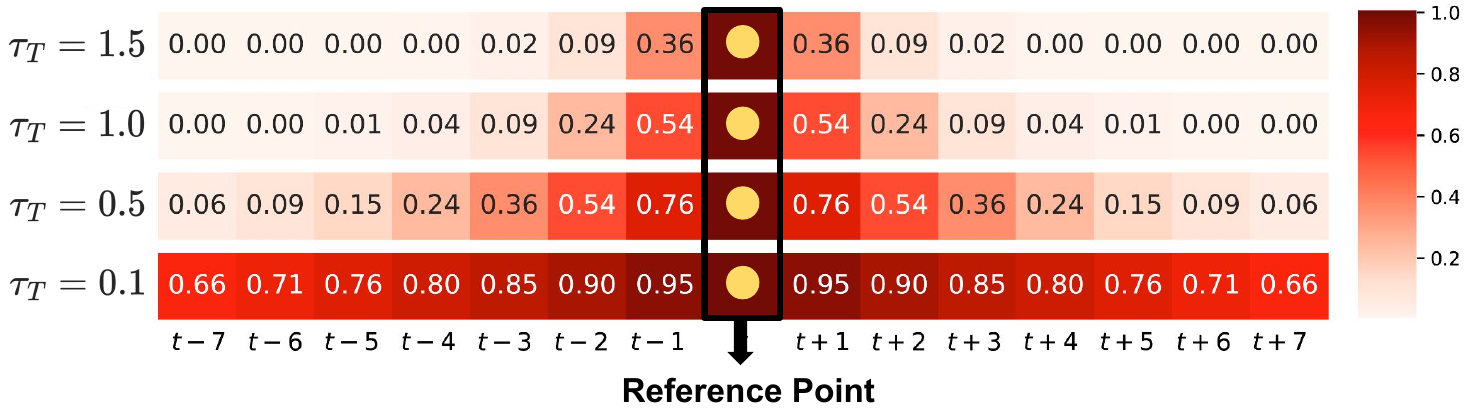}
    \caption{Soft assignments with different $\tau_{T}$.}
    \label{fig:tau}
  \end{subfigure}
  \hfill
  \begin{subfigure}{0.43\textwidth}
    \includegraphics[width=\linewidth]{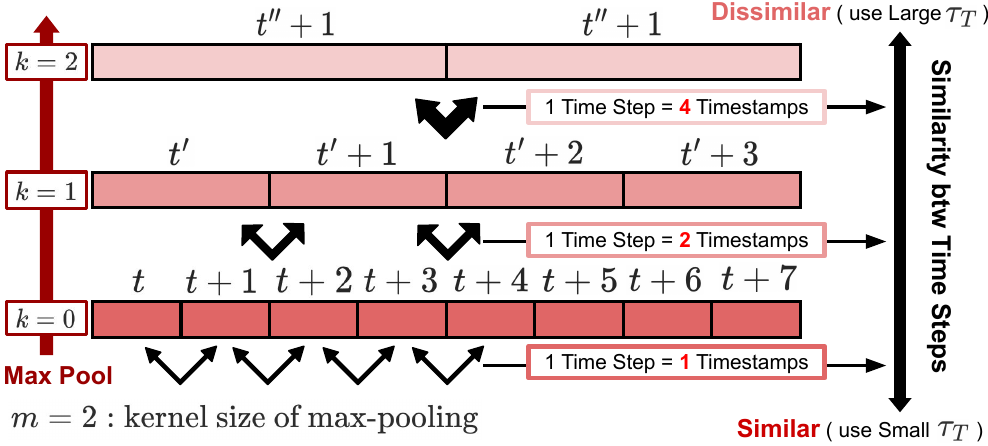}
    \caption{Hierarchical representations.}
    \label{fig:hierarchy}
  \end{subfigure}
  \caption{
  (a) shows examples of soft assignments for soft temporal CL, where a smaller $\tau_T$ results in smoother assignments.
  (b) is an example of hierarchical representations, demonstrating that increasing layer depth results in a larger semantic difference between adjacent time steps, so $\tau_{T}$ should be increased to compensate for it.
  }
  \label{fig:subplots}
  \vspace{-8pt}
\end{figure}

Following the intuition that values in adjacent timestamps are similar,
we propose to compute a soft assignment based on the difference between timestamps for temporal contrastive loss.
Similar to the soft instance-wise contrastive loss, the assignment is
close to one when timestamps get closer and zero when they get farther away.
We define a soft assignment for a pair of timestamps $(t, t^{\prime})$ for the temporal contrastive loss as:
\begin{equation}{
    w_{T} \left(t, t^{\prime}\right)= 2 \cdot \sigma \left( - \tau_{T} \cdot  \left| t-t^{\prime} \right| \right),
    \label{eqn:temp_weight}
}
\end{equation}
where $\tau_{T}$ is a hyperparameter controlling the sharpness.
As the degree of closeness between timestamps varies across datasets,
we tune $\tau_{T}$ to control the degree of soft assignments.
Figure~\ref{fig:tau} illustrates an example of soft assignments with respect to timestamp difference with different $\tau_{T}$. 

\textbf{Hierarchical loss. }
For temporal CL, we consider hierarchical contrasting on intermediate representations in the network $f_{\theta}$ as done in prior CL methods for TS.
Specifically, we adopt the hierarchical contrastive loss proposed in TS2Vec~\citep{yue2022ts2vec},
where the losses are computed on intermediate representations after each max-pooling layer along the temporal axis and then aggregated.
As shown in Figure~\ref{fig:hierarchy}, similarities between adjacent time step decrease after pooling, we adjust $\tau_{T}$  by multiplying $m^k$ in Eq.~\ref{eqn:temp_weight}, i.e., $\tau_{T} = m^k \cdot \tilde{\tau}_{T}$ where $m$ is the kernel size of pooling layers, $k$ is the depth, and $\tilde{\tau}_{T}$ is the base hyperparameter.

Now, let $r_{i,t}=r_{i,t+2T}$ and $\tilde{r}_{i,t} = r_{i,t+T}$ be the embedding vectors from two augmentations of $x_{i}$ at timestamp $t$ for conciseness.
Similar to Eq.~\ref{eqn:inst_prob}, we define a softmax probability of the relative similarity out of all similarities considered when computing the loss as:
\begin{equation}
    p_{T}(i, (t,t^{\prime})) = \frac
    {\exp (r_{i,t} \circ r_{i,t^{\prime}})}
    {\sum_{s=1, s \neq t}^{2T} \exp (r_{i, t} \circ r_{i, s})}.
    \label{eqn:temp_prob}
\end{equation}
Then, the soft temporal contrastive loss for
$x_{i}$ at timestamp $t$ is defined as:
\begin{equation}
{
    \ell_{T}^{(i, t)} = 
    - \log p_{T}(i, (t,t+T)) 
    - \sum_{s=1, s \neq \{t, t+T\}}^{2T}  w_{T}(t,s \text{ mod } T) \cdot \log p_{T}(i, (t,s)).
    \label{eqn:temp_loss}
}
\end{equation}
Similar to the soft instance-wise contrastive loss, this loss can be seen as a generalization of the hard temporal contrastive loss, which is the case when $\forall w_{T}(t,t^{\prime})=0$.

The final loss for SoftCLT is the joint of the soft instance-wise and temporal contrastive losses:
\begin{equation}{
    \mathcal{L}=\frac{1}{4NT} \sum_{i=1}^{2N} \sum_{t=1}^{2T} (\lambda \cdot \ell_{I}^{(i, t)}+ (1-\lambda) \cdot \ell_{T}^{(i, t)}),
}
\end{equation}
where $\lambda$ is a hyperparameter controlling the contribution of each loss, set to 0.5 unless specified.
The proposed loss has an interesting mathematical interpretation that it can be seen as the scaled KL divergence of the softmax probabilities from the normalized soft assignments, where the scale is the sum of soft assignments.
We provide more details in Appendix~\ref{supp_KL}.

\section{Experiments}
We conduct extensive experiments to validate the proposed method and assess its performance in different tasks:
(1)~\textbf{classification} with univariate and multivariate TS,
(2)~\textbf{semi-supervised classification} by (i)~self-supervised learning followed by fine-tuning and (ii)~semi-supervised learning,
(3)~\textbf{transfer learning} in in-domain and cross-domain scenarios, and
(4)~\textbf{anomaly detection} in normal and cold-start settings.
We also conduct ablation studies to validate the effectiveness of SoftCLT as well as its design choices.
Finally, we visualize pairwise distance matrices and t-SNE~\citep{van2008visualizing} of temporal representations to show the effect of SoftCLT over previous methods.
We use the data augmentation strategies of the methods we apply our SoftCLT to: TS2Vec generates two views as TS segments with overlap, and TS-TCC/CA-TCC generate two views with weak and strong augmentations, using the jitter-and-scale and permutation-and-jitter strategies, respectively.

\subsection{Classification}
\begin{figure*}[t]
    \centering
    \begin{minipage}{0.538\textwidth}
        \centering
        \vspace{-5pt}
        \begin{adjustbox}{max width=1.00\textwidth}
        \begin{NiceTabular}{c|cc|cc} 
        \toprule
         & \multicolumn{2}{c}{\text{125 UCR datasets }} & \multicolumn{2}{c}{\text{29 UEA datasets }} \\
        \cmidrule(lr){2-3} \cmidrule(lr){4-5} 
        \text{Method} & \text{Avg. Acc.(\%)} & \text{Avg. Rank} & \text{Avg. Acc.(\%)} & \text{Avg. Rank} \\
        \midrule \text{DTW-D} & 72.7 & 5.30 & 65.0 & 4.60 \\
        \text{TNC} & 76.1 & 4.42 & 67.7 & 4.76 \\
        \text{TST} & 64.1 & 6.19 & 63.5 & 5.26 \\
        \text{TS-TCC} & 75.7 & 4.29 & 68.2 & 4.38 \\
        \text{T-Loss} & 80.6 & 3.50 & 67.5 & 3.86 \\
        \midrule
        \text{TS2Vec} & 83.0 & 2.80 & 71.2 & 3.28 \\
        \text{+ Ours} & \textbf{85.0(+ 2.0)}  & \textbf{1.49} & \textbf{75.1(+ 3.9)} & \textbf{1.86} \\
        \bottomrule
        \end{NiceTabular}
        \end{adjustbox}
        \captionsetup{type=table}
        \caption{Accuracy and rank on UCR/UEA.}
        \label{tbl:cls_accrank}
        \vspace{-10pt}
    \end{minipage}
    \hfill
    \begin{minipage}{0.452\textwidth}
        \centering
        \vspace{-5pt}
        \includegraphics[width=1.00\textwidth]{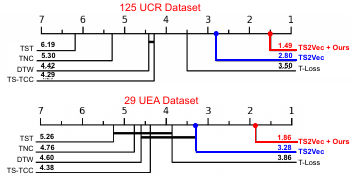} 
        \caption{CD diagram on UCR/UEA.}
        \label{fig:cd_diagram}
        \vspace{-10pt}
    \end{minipage}
\end{figure*}

We conduct experiments on TS classification tasks with 125\footnote{Some of the previous methods cannot handle missing observations, so three of the 128 datasets are omitted.} UCR archive datasets~\citep{dau2019ucr} for univariate TS and 29\footnote{One of the 30 datasets is omitted for a fair comparison with some of the previous methods.} UEA archive datasets~\citep{bagnall2018uea} for multivariate TS, respectively. 
Specifically, we apply SoftCLT to TS2Vec~\citep{yue2022ts2vec}, which has demonstrated SOTA performance on the above datasets.
As baseline methods, we consider DTW-D~\citep{chen2013dtw}, TNC~\citep{tonekaboni2021unsupervised}, TST~\citep{zerveas2021transformer}, TS-TCC~\citep{eldele2021time}, T-Loss~\citep{franceschi2019unsupervised}, and TS2Vec~\citep{yue2022ts2vec}.
The experimental protocol follows that of T-Loss and TS2Vec, where the SVM classifier with the RBF kernel is trained on top of the instance-level representations obtained by max-pooling representations of all timestamps. 
Table~\ref{tbl:cls_accrank} and the critical difference (CD) diagram based on the Wilcoxon-Holm method~\citep{IsmailFawaz2018deep} shown in Figure~\ref{fig:cd_diagram} demonstrate that
the proposed method improves SOTA performance by a significant margin on both datasets in terms of accuracy and rank.
In Figure~\ref{fig:cd_diagram}, 
the best and second-best results for each dataset are in red and blue, respectively.
We also connect methods with a bold line if their difference is not statistically significant in terms of the average rank with a confidence level of 95\%, which shows that
the performance gain by the proposed method is significant.

\subsection{Semi-Supervised Classification}
\begin{table*}[t]
\centering
\begin{adjustbox}{max width=1.00\textwidth}
\begin{NiceTabular}{c|ccc|cc|cc|cccc|cc} 
\toprule
 & \multicolumn{13}{c}{1\% of labeled data} \\
\cmidrule(lr){2-14}
 & \multicolumn{7}{c}{ Self-supervised learning } & \multicolumn{6}{c}{ Semi-supervised learning } \\
\midrule
Dataset & SSL-ECG & CPC & SimCLR & TS2Vec & + Ours & TS-TCC & + Ours & Mean-Teacher & DivideMix & SemiTime & FixMatch & CA-TCC & + Ours \\
\midrule 
HAR & 60.0 / 54.0 & 65.4 / 63.8 & 65.8 / 64.3 & \underline{88.6} / \underline{88.5} & \textbf{91.0} / \textbf{91.0} & 70.5 / 69.5 & 82.9 / 82.8 & 75.9 / 74.0 & 76.5 / 75.4 & \underline{77.6} / \underline{76.3} & 76.4 / 75.6 & 77.3 / 76.2 & \textbf{90.6} / \textbf{90.6} \\
Epilepsy & 89.3 / 86.0 & 88.9 / 85.8 & 88.3 / 84.0 & 95.8 / 93.4 & \textbf{96.3} / \underline{94.1} & 91.2 / 89.2 & \underline{95.6} / \textbf{95.6} & 91.5 / 90.6 & 90.9 / 89.4 & 91.6 / 90.8 & \underline{93.2} / \underline{92.2} & 92.0 / 91.9 & \textbf{97.9} / \textbf{97.9} \\
Wafer & 93.4 / 76.1 & 93.5 / 78.4 & 93.8 / 78.5 & 67.9 / 56.1 & \underline{95.3} / \underline{88.1} & 93.2 / 76.7 & \textbf{96.5} / \textbf{96.5} & 94.7 / 84.7 & 93.2 / 82.0 & 94.4 / 84.4 & 95.0 / 84.8 & \underline{95.1} / \underline{85.1} & \textbf{98.9} / \textbf{98.8} \\
FordA & 67.9 / 66.2 & 75.8 / 75.2 & 55.9 / 55.7 & \underline{86.4} / \underline{86.4} & \textbf{87.1} / \textbf{87.1} &80.6 / 80.0 & 81.5 / 81.2 & 71.7 / 71.5 & 73.7 / 73.3 & 75.1 / 74.4 & 74.5 / 74.3 & \underline{82.3} / \underline{81.7} & \textbf{90.6} / \textbf{90.5} \\
FordB & 64.4 / 60.5 & 66.8 / 65.0 & 50.9 / 49.8 & 65.4 / 65.4 & 67.9 / 67.9 &\textbf{78.6} / \textbf{78.6} & \underline{74.8} / \underline{74.8} & 65.9 / 65.8 & 54.5 / 54.1 & 67.6 / 67.5 & 56.7 / 55.4 & \underline{73.8} / \underline{73.0} & \textbf{78.3} / \textbf{78.2} \\
POC & 62.5 / 41.2 & \underline{64.8} / 48.2 & 61.5 / 38.4 & 63.1 / 62.8 & 63.6 / 62.8 & 63.8 / 48.1 & \textbf{65.4} / \textbf{64.6} & 62.1 / 40.8 & 62.1 / 40.7 & 62.0 / 40.4 & 61.9 / 40.0 & \underline{63.4} / \underline{49.3} & \textbf{73.3} / \textbf{71.7} \\
StarLightCurves & 78.3 / 72.0 & 80.8 / 74.4 & 80.6 / 71.6 & 82.9 / 60.6 & \underline{85.6} / 62.9&\textbf{86.0} / \underline{79.2} & \textbf{86.0} / \textbf{79.3} & 79.4 / 77.7 & 79.0 / 77.2 & 79.5 / \underline{77.8} & 77.2 / 71.6 & \underline{85.8} / \underline{77.8} & \textbf{94.1} / \textbf{94.2} \\
ElectricDevices & 60.1 / 50.0 & 59.3 / 48.9 & 62.5 / 51.2 & 57.6 / 48.6 & 62.0 / 53.0 & \underline{63.6} / \underline{56.4} & \textbf{64.6} / \textbf{63.2} & 48.9 / 48.3 & 59.8 / 49.4 & 57.3 / 48.1 & 58.2 / 46.9 & \underline{65.9} / \underline{56.7} & \textbf{70.3} / \textbf{68.8} \\
\bottomrule 
\toprule 
 & \multicolumn{13}{c}{5\% of labeled data} \\
\cmidrule(lr){2-14}
 & \multicolumn{7}{c}{ Self-supervised learning } & \multicolumn{6}{c}{ Semi-supervised learning } \\
\midrule 
HAR & 63.7 / 58.6 & 75.4 / 74.7 & 75.8 / 74.9 & 91.1 / 91.0 & \underline{92.1} / \underline{92.1} & 77.6 / 76.7 & \textbf{92.6} / \textbf{92.6} & 88.2 / 88.1 & \underline{89.1} / \underline{89.1} & 87.6 / 87.1 & 87.6 / 87.3 & 88.3 / 88.3 & \textbf{91.4} / \textbf{91.4} \\
Epilepsy & 92.8 / 89.0 & 92.8 / 90.2 & 91.3 / 89.2 & 96.0 / 93.6 & \underline{96.7} / \underline{94.9} & 93.1 / \underline{93.7} & \textbf{96.2} / \textbf{96.1} & 94.0 / 93.6 & 93.9 / 93.4 & 94.0 / 93.0 & 93.7 / 92.4 & \underline{94.5} / \underline{94.0} & \textbf{98.0} / \textbf{97.9} \\
Wafer & \underline{94.9} / \underline{84.5} & 92.5 / 79.4 & 94.8 / 83.3 & \underline{98.8} / 96.9 & \textbf{98.8} / \underline{96.8} & 93.2 / 81.2 & 98.2 / \textbf{98.2} & 94.4 / 83.8 & 94.7 / 84.6 & 95.0 / 84.7 & 94.9 / 84.4 & \underline{95.8} / \underline{85.2} & \textbf{98.9} / \textbf{98.8} \\
FordA & 73.6 / 70.7 & 86.5 / 86.5 & 69.6 / 68.9 & 91.2 / 91.2 & 92.5 / 92.5 & 89.9 / 89.9 & \textbf{93.2} / \textbf{93.2} & 82.6 / 82.5 & 84.0 / 83.9 & 83.8 / 83.7 & 83.8 / 83.8 & \underline{90.9} / \underline{90.8} & \textbf{93.3} / \textbf{93.3} \\
FordB & 71.7 / 69.8 & \underline{86.3} / \underline{86.2} & 63.0 / 60.7 & 74.9 / 74.9 & 78.8 / 78.6& 86.1 / 85.9 & \textbf{88.0} / \textbf{88.0} & 64.6 / 62.7 & 60.2 / 57.9 & 65.0 / 62.6 & 62.7 / 60.7 & \underline{88.2} / \underline{88.2} & \textbf{89.4} / \textbf{89.4} \\
POC & 62.9 / 43.3 & 66.9 / 44.3 & 62.7 / 42.4 & \underline{70.4} / \underline{68.0} & \textbf{70.9} / \textbf{69.7} & 62.6 / 42.6 & 69.4 / 66.3 & 62.1 / 41.2 & 62.9 / 45.9 & 62.4 / 41.8 & 63.1 / 43.6 & \underline{66.4} / \underline{52.8} & \textbf{73.1} / \textbf{70.7} \\
StarLightCurves & 82.6 / 74.5 & 89.1 / 84.5 & 84.2 / 74.0 & \underline{90.0} / \underline{87.6} & \textbf{92.3} / \textbf{89.8} & 89.6 / 82.7 & 86.2 / 85.5 & 84.9 / 83.9 & 85.6 / 84.1 & 84.6 / 83.8 & 84.1 / 77.5 & \underline{88.8} / \underline{81.1} & \textbf{94.3} / \textbf{94.2} \\
ElectricDevices & 63.7 / 56.1 & 62.4 / 58.1 & 63.9 / 58.6 & 62.9 / 54.7 & 62.4 / 54.4 & \textbf{65.1} / \underline{59.2} & \textbf{65.1} / \textbf{63.8} & 70.1 / 60.9 & \textbf{72.0} / \underline{62.1} & \underline{71.6} / 61.1 & 62.6 / 55.5 & 66.4 / 59.3 & 70.6 / \textbf{68.9} \\
\bottomrule
\end{NiceTabular}
\end{adjustbox}
\caption{\textbf{Semi-supervised classification results.} The table shows the results of fine-tuning self- and semi-supervised models, with $1\%$ and $5\%$ of labels. \textbf{Best results} across each dataset are in bold, while the \underline{second-best results} are underlined.
The accuracy and MF1 score are reported in order.
}
\label{tbl:semi}
\vspace{-12pt}
\end{table*}     

We conduct experiments on semi-supervised classification tasks by adopting SoftCLT to TS-TCC~\citep{eldele2021time} and its extension CA-TCC~\citep{eldele2022self}, which are the methods that incorporate CL into self- and semi-supervised learning, respectively.
As baseline methods, we consider SSL-ECG~\citep{sarkar2020self}, CPC~\citep{oord2018representation}, SimCLR~\citep{chen2020simple} and TS-TCC~\citep{eldele2021time} for self-supervised learning, and Mean-Teacher~\citep{tarvainen2017mean}, DivideMix~\citep{li2020dividemix}, SemiTime~\citep{fan2021semi}, FixMatch~\citep{sohn2020fixmatch} and CA-TCC~\citep{eldele2022self} for semi-supervised learning.
Note that both TS-TCC and CA-TCC perform instance-wise and temporal contrasting, however, their temporal contrasting is achieved by predicting one view's future from another, which is different from the conventional contrastive loss with positive and negative pairs.
Therefore, we adopt our soft temporal contrastive loss as an additional loss to both methods.
For evaluation, we utilize the same experimental settings and datasets of CA-TCC, which includes eight datasets~\citep{anguita2013public, andrzejak2001indications, dau2019ucr}, six of which are from the UCR archive.
We consider two semi-supervised learning scenarios,
(1)~self-supervised learning with unlabeled data followed by supervised fine-tuning with labeled data and
(2)~semi-supervised learning with both labeled and unlabeled data, following CA-TCC~\citep{eldele2022self}.
Table~\ref{tbl:semi} presents the experimental results with both methods in scenarios with 1\% and 5\% labeled datasets, showing that applying SoftCLT achieves the best overall performance across most of the datasets in both scenarios.

\subsection{Transfer Learning}
\begin{table}[t]
  \centering
  \begin{subtable}[t]{0.88\linewidth}
    \centering
    \begin{adjustbox}{max width=1.00\textwidth}
    \begin{NiceTabular}{c|cccc|cccc|cccc|cccc}
    \toprule 
      & \multicolumn{4}{c}{In-domain transfer learning} & \multicolumn{12}{c}{Cross-domain transfer learning} \\
    \cmidrule(lr){2-5} \cmidrule(lr){6-9} \cmidrule(lr){10-13} \cmidrule(lr){14-17}
     & \multicolumn{4}{c}{SleepEEG $\rightarrow$ Epilepsy} & \multicolumn{4}{c}{SleepEEG $\rightarrow$ FD-B} & \multicolumn{4}{c}{SleepEEG $\rightarrow$ Gesture} & \multicolumn{4}{c}{SleepEEG $\rightarrow$ EMG}  \\
    \cmidrule(lr){2-5} \cmidrule(lr){6-9} \cmidrule(lr){10-13} \cmidrule(lr){14-17}
      & ACC. & PRE. & REC. & $\text{F}_1$ & ACC. & PRE. & REC. & $\text{F}_1$ & ACC. & PRE. & REC. & $\text{F}_1$ & ACC. & PRE. & REC. & $\text{F}_1$  \\
    \midrule 
    TS-SD & 89.52 & 80.18 & 76.47 & 77.67 & 55.66 & 57.10 & 60.54 & 57.03 & 69.22 & 66.98 & 68.67 & 66.56 & 46.06 & 15.45 & 33.33 & 21.11 \\
    TS2Vec & 93.95 & 90.59 & 90.39 & 90.45 & 47.90 & 43.39 & 48.42 & 43.89 & 69.17 & 65.45 & 68.54 & 65.70 & 78.54 & 80.40 & 67.85 & 67.66 \\
    Mixing-Up & 80.21 & 40.11 & 50.00 & 44.51 & 67.89 & 71.46 & 76.13 & 72.73 & 69.33 & 67.19 & 69.33 & 64.97 & 30.24 & 10.99 & 25.83 & 15.41\\ 
    CLOCS & 95.07 & 93.01 & 91.27 & 92.06 & 49.27 & 48.24 & 58.73 & 47.46 & 44.33 & 42.37 & 44.33 & 40.14 & 69.85 & 53.06 & 53.54 & 51.39 \\
    CoST & 88.40 & 88.20 & 72.34 & 76.88 & 47.06 & 38.79 & 38.42 & 34.79 & 68.33 & 65.30 & 68.33 & 66.42 & 53.65 & 49.07 & 42.10 & 35.27 \\
    LaST & 86.46 & 90.77 & 66.35 & 70.67 & 46.67 & 43.90 & 47.71 & 45.17 & 64.17 & 70.36 & 64.17 & 58.76 & 66.34 & 79.34 & 63.33 & 72.55 \\
    TF-C & 94.95 & \underline{94.56} & 89.08 & 91.49 & 69.38 & \underline{75.59} & 72.02 & 74.87 & 76.42 & 77.31 & 74.29 & 75.72 & 81.71 & 72.65 & 81.59 & 76.83 \\
    TST & 80.21 & 40.11 & 50.00 & 44.51 & 46.40 & 41.58 & 45.50 & 41.34 & 69.17 & 66.60 & 69.17 & 66.01 & 46.34 & 15.45 & 33.33 & 21.11 \\
    SimMTM & \underline{95.49} & 93.36 & \underline{92.28} & \underline{92.81} & \underline{69.40} & 74.18 & \underline{76.41} & \underline{75.11} & \underline{80.00} & \underline{79.03} & \underline{80.00} & \underline{78.67} & \underline{97.56} & \underline{98.33} & \underline{98.04} & \underline{98.14}\\ 
    \midrule
    TS-TCC & 92.53 & 94.51 & 81.81 & 86.33 & 54.99 & 52.79 & 63.96 & 54.18 & 71.88 & 71.35 & 71.67 & 69.84 & 78.89 & 58.51 & 63.10 & 59.04 \\
    + Ours & \textbf{97.00} & \textbf{97.07} & \textbf{97.00} & \textbf{96.92} & \textbf{80.45} & \textbf{86.84} & \textbf{85.68} & \textbf{85.48} & \textbf{95.00} & \textbf{95.59} & \textbf{95.00} & \textbf{95.12} & \textbf{100} & \textbf{100} & \textbf{100} & \textbf{100} \\
    \bottomrule
    \end{NiceTabular}
    \end{adjustbox}
    \caption{
    Transfer learning under in- and cross-domain scenarios.
    }
    \vspace{6pt}
    \label{tbl:TL1}
  \end{subtable}
  \\
  \begin{subtable}[t]{0.96\linewidth}
    \centering
    \begin{adjustbox}{max width=1.0\textwidth}
    \begin{NiceTabular}{c|cccccccccccc|c}
    \toprule
     & A $\rightarrow$ B & A $\rightarrow$ C & A $\rightarrow$ D & B $\rightarrow$ A & B $\rightarrow$ C & B $\rightarrow$ D & C $\rightarrow$ A & C $\rightarrow$ B & C $\rightarrow$ D & D $\rightarrow$ A & D $\rightarrow$ B & D $\rightarrow$ C &  Avg\\
    \midrule 
    Supervised & 34.38 & 44.94 & 34.57 & 52.93 & 63.67 & 99.82 & 52.93 & 84.02 & 83.54 & 53.15 & 99.56 & 62.43 & 63.8 \\
    \midrule
    TS-TCC & 43.15 & 51.50 & 42.74 & 47.98 & 70.38 & 99.30 & 38.89 & \textbf{98.31} & \textbf{99.38} & 51.91 & \textbf{99.96} & 70.31 & 67.82 \\
    + Ours & \textbf{76.83} & \textbf{74.35} & \textbf{78.34} & \textbf{53.37} & \textbf{75.11} & \textbf{99.38} & \textbf{53.26} & 85.59 & 86.29 & \textbf{53.30} & 93.55 & \textbf{70.93} & \textbf{75.03} (+7.21\%)\\
    \midrule
    CA-TCC & 44.75 & 52.09 & 45.63 & 46.26 & 71.33 & \textbf{100.0} & 52.71 & \textbf{99.85} & \textbf{99.84} & 46.48 & \textbf{100.0} & 77.01 & 69.66 \\
    + Ours & \textbf{76.85} & \textbf{77.16} & \textbf{79.99} & \textbf{53.26} & \textbf{86.36} & \textbf{100.0} & \textbf{53.23} & 99.67 & 99.01 & \textbf{53.56} & \textbf{100.0} & \textbf{84.93} & \textbf{80.34} (+10.68\%)\\
    \bottomrule
    \end{NiceTabular}
    \end{adjustbox}
    \caption{
    Transfer learning without adaptation under self- and semi-supervised settings
    on the FD datasets. TS-TCC and CA-TCC are used as baselines for self- and semi-supervised learning, respectively.
    }
    \label{tbl:TL2}
    \vspace{-5pt}  
  \end{subtable}
  \caption{Results of transfer learning task on both in- and cross-domain settings.}
  \vspace{-11pt}
\end{table}

We conduct experiments on transfer learning for classification in in-domain and cross-domain settings which are used in previous works~\citep{zhang2022self, eldele2021time, eldele2022self,dong2023simmtm}, by adopting our SoftCLT to TS-TCC and CA-TCC.
As baseline methods, we consider TS-SD~\citep{shi2021self}, TS2Vec~\citep{yue2022ts2vec}, Mixing-Up~\citep{wickstrom2022mixing}, CLOCS~\citep{kiyasseh2021clocs}, CoST~\citep{woo2022cost}, LaST~\citep{wang2022learning}, TF-C~\citep{zhang2022self}, TS-TCC~\citep{eldele2021time}, TST~\citep{zerveas2021transformer} and SimMTM~\citep{dong2023simmtm}.
In in-domain transfer learning, the model is pretrained on SleepEEG~\citep{kemp2000analysis} and fine-tuned on Epilepsy~\citep{andrzejak2001indications}, where they are both EEG datasets and hence considered to be in a similar domain.
In cross-domain transfer learning, which involves pretraining on one dataset and fine-tuning on different datasets, the model is pretrained on SleepEEG, and fine-tuned on three datasets from different domains, FD-B~\citep{lessmeier2016condition}, Gesture~\citep{liu2009uwave}, and EMG~\citep{goldberger2000physiobank}. 
Also, we perform transfer learning without adaptation under self-and semi- supervised settings, where source and target datasets share the same set of classes but only 1\% of labels are available for the source dataset, and no further training on the target dataset is allowed.
Specifically, models are trained on one of the four conditions (A,B,C,D) in the Fault Diagnosis (FD) datasets~\citep{lessmeier2016condition} and test on another.
Table~\ref{tbl:TL1} shows the results of both in- and cross-domain transfer learning, and Table~\ref{tbl:TL2} shows the results of both self- and semi-supervised settings with FD datasets.
Notably, SoftCLT applied to CA-TCC improves average accuracy of twelve transfer learning scenarios with FD datasets by 10.68\%.

\subsection{Anomaly Detection}
We conduct experiments on univariate TS anomaly detection (AD) task by adopting SoftCLT to TS2Vec~\citep{yue2022ts2vec} under two different settings:
the normal setting splits each dataset into two halves according to the time order and use them for training and evaluation, respectively, and the cold-start setting pretrains models on the FordA dataset in the UCR archive and evaluates on each dataset.
As baseline methods, we consider SPOT~\citep{siffer2017anomaly}, DSPOT~\citep{siffer2017anomaly}, DONUT~\citep{xu2018unsupervised}, SR~\citep{ren2019time}, for the normal setting, and
FFT~\citep{rasheed2009fourier}, Twitter-AD~\citep{vallis2014novel}, Luminol~\citep{luminol} for the cold-start setting, and TS2Vec~\citep{yue2022ts2vec} for both.
The anomaly score is computed by the L1 distance of two representations encoded from masked and unmasked inputs following TS2Vec.
We evaluate the compared method on
the Yahoo~\citep{laptev2015benchmark} and KPI~\citep{ren2019time} datasets.
We found that suppressing instance-wise CL leads to better AD performance on average, so we report TS2Vec and SoftCLT performances without instance-wise CL; more details can be found in the Appendix~\ref{supp_AD}.
As shown in Table~\ref{tbl:AD}, SoftCLT outperforms the baselines in both settings in terms of the F1 score, precision, and recall.
Specifically, SoftCLT applied to TS2Vec improves the F1 score approximately 2\% in both datasets under both normal and cold-start settings.

 \begin{table}[t]
    \begin{threeparttable}
    \centering
    \begin{subtable}[t]{0.41\textwidth}
        \centering
         \begin{adjustbox}{max width=1.00\textwidth}
          \begin{NiceTabular}{c|ccc|ccc} 
          \toprule
           & \multicolumn{3}{c}{\text{Yahoo}} & \multicolumn{3}{c}{\text{KPI}} \\
          \cmidrule(lr){2-4} \cmidrule(lr){5-7} & $\text{F}_1$ & Prec. & Rec. & $\text{F}_1$ & Prec. & Rec. \\
          \midrule 
          \text{SPOT} & 33.8 & 26.9 & 45.4 & 21.7 & 78.6 & 12.6 \\
          \text{DSPOT} & 31.6 & 24.1 & 45.8 & 52.1 & 62.3 & 44.7 \\
          \text{DONUT} & 2.6 & 1.3 & 82.5 & 34.7 & 37.1 & 32.6 \\
          \text{SR} & 5.63 & 45.1 & 74.7 & 62.2 & 64.7 & 59.8 \\
          \midrule 
          $\text{TS2Vec}^{\ast}$  & \underline{72.3} & 69.3 & 75.7 & \underline{67.6} & 91.1& 53.7 \\
          $\text{+ Ours}$ & \textbf{74.2} & 72.2 & 76.5 & \textbf{70.1} & 91.6 & 57.0 \\
          \bottomrule
          \end{NiceTabular}
        \end{adjustbox}
        \caption{Results of AD task on normal setting.}
        \label{tbl:ad_normal}
    \end{subtable}
    \hspace{0.5cm}
    \begin{subtable}[t]{0.41\textwidth}
        \centering
        \begin{adjustbox}{max width=1.00\textwidth}
        \begin{NiceTabular}{c|ccc|ccc} 
        \toprule
        & \multicolumn{3}{c}{\text{Yahoo}} & \multicolumn{3}{c}{\text{KPI}} \\
        \cmidrule(lr){2-4} \cmidrule(lr){5-7} & $\text{F}_1$ & \text{Prec.} & \text{Rec.} & $\text{F}_1$ & \text{Prec.} & \text{Rec.} \\
        \midrule
        \text{FFT} & 29.1 & 20.2 & 51.7 & 53.8 & 47.8 & 61.5 \\
        \text{Twitter-AD} & 24.5 & 16.6 & 46.2 & 33.0 & 41.1 & 27.6 \\
        \text{Luminol} & 38.8 & 25.4 & 81.8 & 41.7 & 30.6 & 65.0 \\
        \text{SR} & 52.9 & 40.4 & 76.5 & 66.6 & 63.7 & 69.7 \\
        \midrule
        $\text{TS2Vec}^{\ast}$  & \underline{74.0} & 70.7 & 77.6 & \underline{68.9} & 89.3 & 56.2 \\
        $\text{+ Ours}$ & \textbf{76.2} & 75.3 & 77.3 & \textbf{70.7} & 92.1 & 57.4 \\
        \bottomrule
        \end{NiceTabular}
      \end{adjustbox}
        \caption{Results of AD task on cold-start setting.}
        \label{tbl:ad_coldstart}
    \end{subtable}
    \begin{flushleft}
        \par\noindent\hspace{10pt}\rule{.4\textwidth}{0.5pt}
    \end{flushleft}
    \begin{tablenotes}
    \begin{scriptsize}
    \item[$\ast$] We used the official code to replicate the results without the instance-wise contrastive loss.
    \end{scriptsize}
    \end{tablenotes}
    \end{threeparttable}
    \vspace{0cm}
    \caption{
    Anomaly detection results.
    }
    \label{tbl:AD}
\end{table}

\subsection{Ablation Study}
\begin{table*}[t]
    \centering
    \begin{subtable}[h]{0.38\textwidth}
        \centering
        \begin{adjustbox}{max width=\textwidth}
        \begin{NiceTabular}{cc|c|c} \toprule
        \multicolumn{2}{c}{Soft assignment} & UCR datasets & UEA datasets \\  
        \cmidrule(lr){1-2} \cmidrule(lr){3-3} \cmidrule(lr){4-4}
        Instance-wise & Temporal & Avg. Acc.(\%) & Avg. Acc.(\%) \\
        \midrule
          &  & 82.3 & 70.5\\
          \cmark &  & 83.9 (+1.6) & 73.0 (+2.5) \\
          & \cmark & 83.7 (+1.4) & 73.8 (+3.3) \\
         \cmark & \cmark & \textbf{85.0} (+2.7) & \textbf{74.2} (+3.7) \\
        \bottomrule
        \end{NiceTabular}
        \end{adjustbox}
        \caption{Application of soft assignments.}
        \label{tbl:ablation1}
    \end{subtable}
    \hspace{0.05cm}
    \begin{subtable}[h]{0.185\textwidth}
        \centering
        \begin{adjustbox}{max width=\textwidth}
            \begin{NiceTabular}{c|c}
                \toprule
                \multicolumn{2}{c}{Temporal CL} \\
                \midrule
                Method & Avg. Acc.(\%) \\
                \midrule
                Neighbor & 76.1\\
                Linear & 77.2\\
                Gaussian & 83.5\\
                Sigmoid & \textbf{83.7}\\
                \bottomrule
            \end{NiceTabular}
        \end{adjustbox}
        \caption{Assignment func.}
        \label{tbl:ablation2}
    \end{subtable}
    \hspace{0.05cm}
    \begin{subtable}[h]{0.155\textwidth}
        \centering
        \begin{adjustbox}{max width=\textwidth}
            \begin{NiceTabular}{c|c} 
                \toprule
                \multicolumn{2}{c}{Instance-wise CL} \\
                \midrule
                $\alpha$ & Avg. Acc.(\%) \\  
                \midrule 
                0.25 & 83.0 \\
                0.50 & \textbf{83.9} \\
                0.75 & 83.4 \\
                1.00 & 83.1 \\
                \bottomrule
            \end{NiceTabular}
        \end{adjustbox}
        \caption{Upper bound.}
        \label{tbl:ablation3}
    \end{subtable}
    \hspace{0.05cm}
    \begin{subtable}[h]{0.175\textwidth}
        \centering
        \begin{adjustbox}{max width=\textwidth}
            \begin{NiceTabular}{c|cc}
                \toprule
                Inst. CL & \multicolumn{2}{c}{Temporal CL} \\
                \cmidrule(lr){1-1} \cmidrule(lr){2-3}
                Metric & Hard & Soft \\
                \midrule
                COS & 83.7 & 84.7 \\ 
                EUC & 83.9 & 84.8 \\ 
                DTW & 83.9 & \textbf{85.0} \\ 
                TAM & 83.9 & \textbf{85.0} \\ 
                \bottomrule
            \end{NiceTabular}
        \end{adjustbox}
        \caption{Distance func.}
        \label{tbl:ablation4}
    \end{subtable}
    \caption{
    Ablation study results.
    }
    \label{tbl:ablation}
    \vspace{-12pt}
\end{table*}

\textbf{Effectiveness of SoftCLT. }
Table~\ref{tbl:ablation1} shows the effect of soft assignments from the standard hard CL.
Applying soft assignments to instance-wise or temporal CL provides a performance gain, and applying them to both dimensions results in the best performance, improving the accuracy on the UCR and UEA datasets by 2.7\% and 3.7\%, respectively.

\setlength{\intextsep}{5pt} 
\setlength{\columnsep}{5pt} 

\textbf{Design choices for soft temporal CL. }
Table~\ref{tbl:ablation2} compares different choices of the soft assignment $w_{T}$.
\textbf{Neighbor} takes neighborhood within a window around the reference point as positive and the others as negative.
\textbf{Linear} gives soft assignments linearly proportional to the time difference from the reference point, where the most distant one gets the value of zero.
\textbf{Gaussian} gives soft assignments based on a Gaussian distribution with the mean of the reference point and the standard deviation as a hyperparameter.
Among them, \textbf{Sigmoid} in Eq.~\ref{eqn:temp_weight} shows the best performance as shown in Table~\ref{tbl:ablation2}.

\textbf{Upper bound for soft instance-wise CL. }
In the soft instance-wise contrastive loss, $\alpha$ is introduced to avoid giving the same assignment to pairs of the same TS and pairs of the different TS with the distance of zero,
where $\alpha=1$ makes both cases to have the same assignment.
Table~\ref{tbl:ablation3} studies the effect of tuning $\alpha$.
Based on the results, $\alpha=0.5$ is the best, i.e., the similarity of the pairs of the same TS should be strictly larger than other pairs, but not by much.

\textbf{Distance metrics for soft instance-wise CL. }
Table~\ref{tbl:ablation4} compares different choices of the distance metric $D$ in Eq.~\ref{eqn:inst_weight}:
cosine distance (COS), Euclidean distance (EUC), dynamic time warping (DTW), and time alignment measurement (TAM)~\citep{folgado2018time} on 128 UCR datasets,
where the baseline is TS2Vec and the hard or best soft temporal CL is applied together.
The result shows that the improvement by soft instance-wise CL is robust to the choice of the distance metric.
We use DTW throughout all other experiments because DTW is well-studied, commonly used in the literature and fast algorithms such as FastDTW are available.

\subsection{Analysis}
\begin{figure*}[t]
    \centering
    \begin{minipage}{0.49\textwidth}
        \centering
        \begin{adjustbox}{max width=\textwidth}
            \begin{NiceTabular}{c|c|cc|c} 
                \toprule
                \multirow{2}{*}{Method} & \multirow{2}{*}{Total} & \multicolumn{2}{c}{Length of time series} & \multirow{2}{*}{Gap (A-B)}\\
                \cmidrule(lr){3-4}
                 & & $\leq 200$ (A) & $> 200$ (B) &  \\
                \midrule
                TS2Vec & 82.3 & 88.1 & 79.6 & 5.8 \\
                \midrule
                + NNCLR & 66.0 & 82.6 & 58.2 & 24.4 \\
                + ASCL & 76.5 & 86.6 & 71.8 & 14.8 \\
                \midrule 
                + Ours & 85.0 & 89.8 & 81.9 & 7.9 \\
                \bottomrule
            \end{NiceTabular}
        \end{adjustbox}
        \captionsetup{type=table}
        \caption{
        Comparison of soft CL methods.
        }
    \label{tbl:soft-assignments}
    \end{minipage}
    \hfill
    \begin{minipage}{0.49\textwidth}
               \centering
        \begin{adjustbox}{max width=1.00\textwidth}
    \begin{NiceTabular}{c|cc}
        \toprule
        Temporal CL & \multicolumn{2}{c}{Seasonality} \\
        \cmidrule(lr){1-1} \cmidrule(lr){2-3}
        Soft & Low (103/128) & High (25/128) \\ 
        \midrule
        \xmark & 84.1 & 80.1 \\
        \cmark & 85.6 & 81.7 \\
        \midrule
        Gain & +1.5 & +1.6 \\
        \bottomrule
    \end{NiceTabular}
    \end{adjustbox}
        \captionsetup{type=table}
        \caption{
        Effect of soft temporal CL by seasonality.
        }
 \label{tbl:season_trend}
    \end{minipage}
    \vspace{-6pt}
\end{figure*}

\begin{figure*}[t]
    \centering
    \begin{minipage}{0.47\textwidth}
        \centering
        \includegraphics[width=\linewidth]{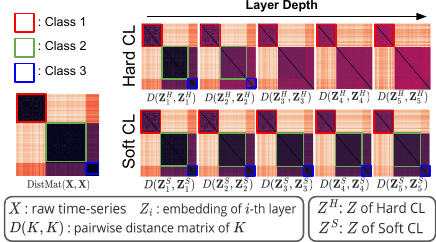} 
        \caption{Pairwise distance matrices.
        }
    \label{fig:dtw_change}
    \end{minipage}
    \vspace{-4pt}
    \hfill
    \begin{minipage}{0.51\textwidth}
        \centering
        \includegraphics[width=\linewidth]{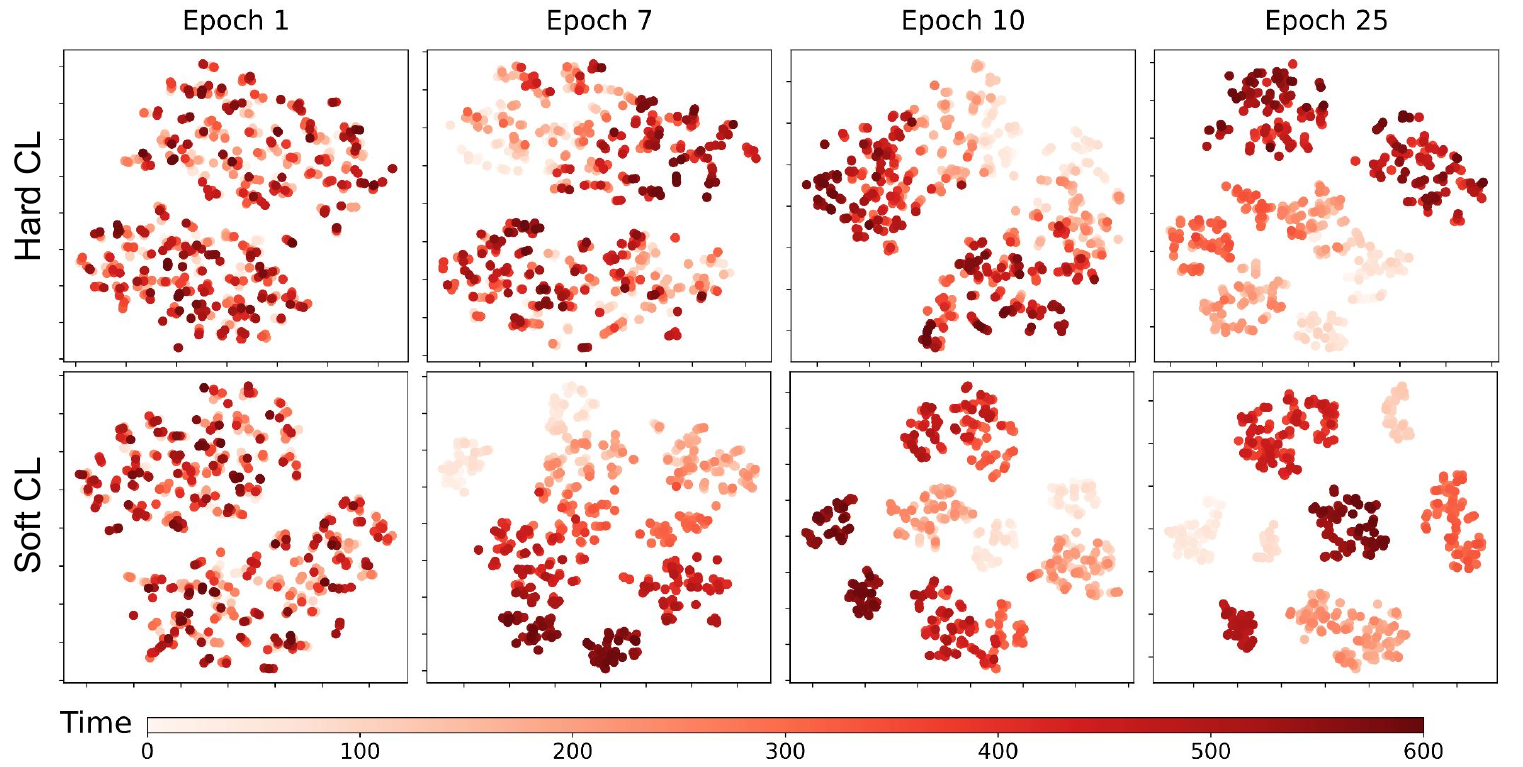} 
        \caption{Visualization of temporal representations.}
    \label{fig:clustering_temporal}
    \end{minipage}
    \vspace{-4pt}
\end{figure*}

\textbf{Comparison with soft CL methods in computer vision. } 
While soft CL methods have been proposed in other domains, they compute soft assignments on the embedding space because it is difficult to measure the similarities on the data space, particularly in computer vision.
However, we argue that the similarities on the data space is indeed a strong self-supervision, leading to better representation learning.
To confirm this, we compare SoftCLT with 
soft CL methods proposed in other domains working on the embedding space:
NNCLR~\citep{dwibedi2021little} and ASCL~\citep{feng2022adaptive}, on UCR datasets. 
For a fair comparison, we apply all compared methods to TS2Vec under the same setting.
As shown in Table~\ref{tbl:soft-assignments}, different from the proposed method, NNCLR and ASCL deteriorate the performance of TS2Vec, implying that similarities measured on the data space is strong self-supervision, while similarities measured on the learnable embedding space might not be useful in some domains.
To further investigate the failure modes of the previous methods, we categorize datasets by the average TS length of 200 in Table~\ref{tbl:soft-assignments}, and observe that previous methods fail to capture the similarities of long TS data.
    
\textbf{Robustness to seasonality. }
An assumption behind the proposed soft temporal CL is that
values in adjacent timestamps are similar, 
which may raise a concern that seasonality in TS might not be captured.
To address this, we categorize UCR datasets based on seasonality by ADF test \citep{sims1990inference} at the significance level of $p=0.05$.
As shown in Table~\ref{tbl:season_trend}, the performance gain by SoftCLT is consistent regardless of the seasonality.
Our conjecture is that TS in the real world usually do not exhibit the perfect seasonality, as indicated by the ADF test result, such that SoftCLT takes advantage of the non-seasonal portions.
Meanwhile, previous works have tried to decompose trend and seasonality in TS for representation learning~\citep{wang2022learning, woo2022cost}.
However, this may not be realistic for TS that are neither simultaneously auto-regressive nor stationary~\citep{shen2022respecting}.
In summary, we do not consider seasonality in TS directly, because it is not only challenging to extract but we can still achieve good performance without considering it in practice.

\textbf{Instance-wise relationships. }
To see whether instance-wise relationships are preserved
in the encoder,
we visualize the pairwise instance-wise distance matrices of representations on the InsectEPGRegularTrain dataset from the UCR archive~\citep{dau2019ucr} extracted from each layer, where the brighter color indicates the lower distance between instances.
The top and bottom panels of Figure~\ref{fig:dtw_change} show the changes in pairwise distance matrices of representations
as depth progresses when adopting hard and soft CL, respectively. 
The results indicate that
SoftCLT preserves the relationships between TS instances throughout encoding, while the standard hard CL fails to preserve them.

\textbf{Temporal relationships. }
To assess the quality of temporal relationships captured by SoftCLT, we apply t-SNE~\citep{van2008visualizing} to visualize the temporal representations, which are representations of each timestamp in a single TS.
Figure~\ref{fig:clustering_temporal} compares t-SNE of the representations learned with hard and soft CL over different training epochs, with the points getting darker as time progresses. 
While hard CL finds coarse-grained neighborhood relationships such that it fails to distinguish late timestamps in dark red, soft CL finds more fine-grained relationships.

\section{Conclusion}
In this paper, we present a soft contrastive learning framework for time series. 
In contrast to previous methods that give hard assignments to sample pairs, our approach gives soft assignments based on the instance-wise and temporal relationships on the data space.
We demonstrate the effectiveness of our method in a range of tasks, leading to significant improvements in performance.
We hope our work enlightens the effectiveness of self-supervision from the data space and motivates future works on contrastive representation learning in various domains to take account of it.

\newpage
\section*{Ethics Statement}
The proposed soft contrastive learning algorithm for time series has a potential to make a significant impact on the field of representation learning for time series data. 
The ability to apply this algorithm to various tasks and solve the general problem of time series representation learning is promising.
In particular, the algorithm can be applied to transfer learning, which may be useful in scenarios with small datasets for downstream tasks.
Furthermore, we expect that the idea of utilizing self-supervision from the data space for contrastive representation learning motivates future works in various domains.

However, as with any algorithm, there are ethical concerns to be considered.
One potential ethical concern is a potential for the algorithm to perpetuate biases that may exist in the datasets used for pretraining. 
For example, if the pretraining dataset is imbalanced with respect to certain demographic attributes, this bias may be transferred to fine-tuning, potentially leading to biased predictions.
It is essential to evaluate and address potential biases in the pretraining dataset before using the algorithm in real-world scenarios.

To ensure responsible use of the algorithm, we will make the datasets and code publicly available. 
Public availability of datasets and code allows for transparency and reproducibility, allowing other researchers to evaluate and address potential biases and misuse. 

\section*{Acknowledgements}
This work was supported by the National Research Foundation of Korea (NRF) grant funded by the Korea government (MSIT) (2020R1A2C1A01005949, 2022R1A4A1033384, RS-2023-00217705), the MSIT(Ministry of Science and ICT), Korea, under the ICAN(ICT Challenge and Advanced Network of HRD) support program (RS-2023-00259934) supervised by the IITP(Institute for Information \& Communications Technology Planning \& Evaluation), the Yonsei University Research Fund (2023-22-0071), and the Son Jiho Research Grant of Yonsei University (2023-22-0006).

\bibliography{iclr2024_conference.bib}
\bibliographystyle{iclr2024_conference}

\newpage
\appendix
\input{camera_ready_appendix.tex}

\end{document}

%% file: math_commands.tex
\usepackage{amsmath,amsfonts,bm}

\def\eqref#1{equation~\ref{#1}}

\def\1{\bm{1}}

\DeclareMathAlphabet{\mathsfit}{\encodingdefault}{\sfdefault}{m}{sl}
\SetMathAlphabet{\mathsfit}{bold}{\encodingdefault}{\sfdefault}{bx}{n}



%% file: camera_ready_appendix.tex
\appendix
\numberwithin{table}{section}
\numberwithin{figure}{section}
\numberwithin{equation}{section}

\section{Dataset Description}
\subsection{Classification}
For time series classification, we use the UCR archive~\citep{dau2019ucr} and UEA archive~\citep{bagnall2018uea}.
The UCR archive contains 128 univariate datasets, while the UEA archive contains 30 multivariate datasets.
Among them,
some datasets cannot be handled by T-Loss~\citep{franceschi2019unsupervised}, TS-TCC~\citep{eldele2021time}, and TNC~\citep{tonekaboni2021unsupervised} due to missing observations, such as DodgerLoopDay, DodgerLoopGame, and DodgerLoopWeekend. Additionally, there is no reported result for the DTW-D~\citep{chen2013dtw} on the InsectWingbeat dataset in the UEA archive. Hence, the comparison is conducted using the remaining 125 UCR datasets and 29 UEA datasets in the main paper.
However, TS2Vec works well on all UCR and UEA datasets, so we experiment with all 128 UR datasets and 30 UEA datasets for ablation studies
for our method on top of TS2Vec.

\subsection{Semi-supervised Classification}
Table~\ref{tbl:datastat_semi} describes the summary of the statistical information for eight datasets~\citep{anguita2013public, andrzejak2001indications, dau2019ucr} used in semi-supervised classifiaction, including the number of training and testing samples, data length, the number of sensor channels, and the number of classes.

\begin{table*}[h]
\centering
\captionsetup{justification=centering}
\begin{adjustbox}{max width=0.80\textwidth}
\begin{NiceTabular}{c|ccccc}
\toprule 
\text { Dataset } & \text { \# Train } & \text { \# Test } & \text { Length } & \text { \# Channel } & \text { \# Class } \\
\midrule 
\text { HAR } & 7,352 & 2,947 & 128 & 9 & 6 \\
\text { Epilepsy } & 9,200 & 2,300 & 178 & 1 & 2 \\
\text { Wafer } & 1,000 & 6,174 & 152 & 1 & 2 \\
\text { FordA } & 1,320 & 3,601 & 500 & 1 & 2 \\
\text { FordB } & 3,636 & 810 & 500 & 1 & 2 \\
\text { POC } & 1,800 & 858 & 80 & 1 & 2 \\
\text { StarLightCurves } & 1,000 & 8,236 & 1,024 & 1 & 3 \\
\text { ElectricDevices } & 8,926 & 7,711 & 96 & 1 & 7 \\
\bottomrule
\end{NiceTabular}
\end{adjustbox}
\centering
\caption{Eight datasets used for semi-supervised classification} 
\label{tbl:datastat_semi}
\end{table*}

\subsection{Transfer Learning}
We evaluate our approach on various datasets, which cover a wide range of application scenarios, including neurological healthcare, human activity recognition, mechanical fault detection, and physical status monitoring. Table~\ref{datastat2} describes the datasets for in-domain and cross-domain transfer learning. Fault Diagnosis (FD) datasets were used for transfer learning under self- and semi-supervised settings. The data statistics are described below.

\begin{table*}[h]
\begin{adjustbox}{max width=0.99\textwidth}
    \begin{NiceTabular}{cc|cccccc}
    \toprule 
     & & Dataset & \# Samples & \# Channels & \# Classes & Length & Freq (Hz)\\
    \midrule 
    \multicolumn{2}{c}{Pre-training} & SleepEEG & 371,055 & 1 & 5 & 200 & 100 \\
    \midrule
    \multirow{4}{*}{Fine-tuning} & In-domain & Epilepsy & 60 / 20 / 11,420 & 1 & 2 & 178 & 174 \\
    \cmidrule(l{0pt}r{0pt}){2-8}
    & \multirow{3}{*}{Cross-domain} & FD-B & 60 / 21 / 13,559 & 1 & 3 & 5,120 & 64,000 \\
    &  & Gesture & 320 / 120 / 120 & 3 & 8 & 315 & 100 \\    
    &  & EMG & 122 / 41 / 41 & 1 & 3 & 1,500 & 4,000 \\
    \bottomrule
    \end{NiceTabular}
\end{adjustbox}
\caption{
In the four application scenarios, we utilize a pre-training dataset and a fine-tuning dataset, with the latter having a sample size denoted by "A/B/C," where each denotes the number of samples used for fine-tuning, validation, and testing, respectively. 
Our evaluation also focuses on small datasets, with a very limited sample size of less than 320 samples in the fine-tuning dataset, ensuring that the fine-tuning set is balanced in terms of classes. 
This approach enables us to test our model's effectiveness on small datasets, which has practical significance.}
\label{datastat2}
\end{table*}  
(1) \textbf{SleepEEG}~\citep{kemp2000analysis} dataset contains EEG recordings of 153 whole-night sleep sessions from 82 healthy individuals. We segmented the EEG signals using a non-overlapping approach, following the same preprocessing method as (Zhang et al., 2022), to obtain 371,055 univariate brainwaves, each sampled at 100 Hz and categorized into one of five sleep stages: Wake, Non-rapid eye movement (3 sub-states), and Rapid Eye Movement. When using SleepEEG dataset as a source dataset in transfer learning task, we use cosine similarity instead of DTW due to the property of EEG datasets~\citep{li2022shveegc}.

(2) \textbf{Epilepsy}~\citep{andrzejak2001indications} dataset monitors brain activity using a single-channel EEG sensor on 500 subjects, with each subject being recorded for 23.6 seconds. The dataset is sampled at 178 Hz and contains 11,500 samples. We followed the same preprocessing method as (Zhang et al., 2022) and classified the first four classes (eyes open, eyes closed, EEG measured in the healthy brain region, and EEG measured in the tumor region) of each sample as positive, while the remaining classes (whether the subject has a seizure episode) were classified as negative.

(3) \textbf{FD-B}~\citep{lessmeier2016condition} dataset is collected from electromechanical drive systems and monitors the condition of rolling bearings to detect their failures based on monitoring conditions such as speed, load torque, and radial force. It consists of 13,640 samples, each recorded at 64k Hz and categorized into three classes: undamaged, inner damaged, and outer damaged.

(4) \textbf{Gesture}~\citep{liu2009uwave} dataset includes data on 8 hand gestures based on hand movement paths recorded by an accelerometer. The eight gestures are hand swiping left, right, up, and down, hand waving in a counterclockwise or clockwise circle, hand waving in a square, and waving a right arrow. The dataset contains 440 balanced classification labels, with each sample having eight different categories of gestures.

(5) \textbf{EMG}~\citep{goldberger2000physiobank} dataset consists of 163 single-channel EMG recordings from the tibialis anterior muscle of three healthy volunteers suffering from neuropathy and myopathy. Each sample is associated with one of three classes, with each class representing a different patient. The dataset is sampled at 4K Hz.

(6) \textbf{FD}~\citep{lessmeier2016condition} dataset was obtained by monitoring the sensor readings of a bearing machine while it operated under four distinct working conditions. Each working condition can be regarded as a separate domain since they exhibit unique features, such as variations in rotational speed and load torque. Within each domain, there are three categories: two fault classes, inner and outer fault, and one healthy class. The FD dataset has 8,184 training samples, 2,728 test samples, a data length of 5,120, one channel, and three classes. Our main goal is to use this dataset to conduct transferability experiments under both self- and semi-supervised settings and demonstrate the efficiency of our approach in transfer learning situations.

\subsection{Anomaly Detection}
We employed Yahoo~\citep{laptev2015benchmark} and KPI~\citep{ren2019time} for the anomaly detection task. 
Yahoo is a benchmark dataset that contains 367 hourly sampled time series with annotated anomaly points. 
This dataset covers a wide range of anomaly types, including outliers and change-points. 
KPI is a competition dataset released by AIOPS Challenge in 2019. 
It contains several minutely sampled real KPI curves from diverse internet companies. 

\section{Baseline Methods}
\paragraph{Classification: }
The results of all baseline methods for the classification task (DTW-D~\citep{chen2013dtw}, TNC~\citep{tonekaboni2021unsupervised}, TST~\citep{zerveas2021transformer}, TS-TCC~\citep{eldele2021time}, T-Loss~\citep{franceschi2019unsupervised}, and TS2Vec~\citep{yue2022ts2vec}) are reported in \citet{yue2022ts2vec}.
\begin{itemize}[leftmargin=22pt, itemsep=4pt]
    \item DTW-D~\citep{chen2013dtw}: DTW-D (Dynamic Time Warping-Delta) is a variant of DTW under semi-supervised learning settings.
    \item TNC~\citep{tonekaboni2021unsupervised}: TNC (Temporal Neighborhood Coding) defines temporal neighborhood of window using normal distribution, and defines samples in neighborhood and non-neighborhood as positives and negatives, respectively.
    \item TST~\citep{zerveas2021transformer}: TST (Time Series Transformer) adopts the masked modeling paradigm to time series domain, where the goal is to reconstruct the masked time stamps. 
    \item TS-TCC~\citep{eldele2021time}: TS-TCC (Time-Series representation learning framework via Temporal and Contextual Contrasting) proposes a new temporal contrastive loss by making the augmentations predict each other's future.
    \item T-Loss~\citep{franceschi2019unsupervised}: T-Loss is a triplet loss designed for time series. It samples a random subseries from a time series and treats them as positive when they belong to its subseries, and negative if belong to subseries of other time series.
    \item TS2Vec~\citep{yue2022ts2vec}: TS2Vec splits time series into several subseries and defines hierarchical contrastive loss in both instance-wise and temporal dimensions.
\end{itemize}

\paragraph{Semi-supervised classification: }
The results of all baseline methods for semi-supervised classification using self-supervised methods (SSL-ECG~\citep{sarkar2020self}, CPC~\citep{oord2018representation}, SimCLR~\citep{chen2020simple}, TS-TCC~\citep{eldele2021time}) and semi-supervised methods (Mean-Teacher~\citep{tarvainen2017mean}, DivideMix~\citep{li2020dividemix}, SemiTime~\citep{fan2021semi}, FixMatch~\citep{sohn2020fixmatch}, CA-TCC~\citep{eldele2022self}) are reported in \citet{eldele2022self}.
\begin{itemize}[leftmargin=22pt, itemsep=4pt]
    \item SSL-ECG~\citep{sarkar2020self}: SSL-ECG (Self-supervised ECG Representation Learning for Emotion Recognition) proposes ECG-based emotion recognition using multi-task self-supervised learning
    \item CPC~\citep{oord2018representation}: CPC (Contrastive Predictive Coding) combines predicting future observations (predictive coding) with a probabilistic contrastive loss.
    \item SimCLR~\citep{chen2020simple}: SimCLR proposes a simple framework for contrastive learning of visual representations, without requiring specialized architectures or a memory bank.
    \item Mean-Teacher~\citep{tarvainen2017mean}: Mean-Teacher is an algorithm for semi-supervised algorithm, that averages model weights instead of predictions.
    \item DivideMix~\citep{li2020dividemix}: DivideMix uses a mixture model to divide training data into labeled clean samples and unlabeled noisy samples, and trains a model on both sets in a semi-supervised way.
    \item SemiTime~\citep{fan2021semi}: SemiTime conducts supervised classification on labeled time series data and self-supervised prediction of temporal relations on unlabeled time series data. It achieves this by sampling segments of past-future pairs from the same or different candidates and training the model to distinguish between positive and negative temporal relations between those segments.
    \item FixMatch~\citep{sohn2020fixmatch}: FixMatch generates pseudo-labels using the model's predictions on weakly-augmented unlabeled images, and retain the pseudo-label with a high-confidence prediction. Then, the model is trained to predict the pseudo-label when fed a strongly-augmented version of the same image.
    \item CA-TCC~\citep{eldele2022self}: CA-TCC (Self-supervised Contrastive Representation Learning for Semi-supervised Time-Series Classification) is the extension of TS-TCC to the semi-supervised settings, and adopts the same contrastive loss as TS-TCC.
\end{itemize}

\paragraph{Transfer learning: }
The results of baseline methods for transfer learning in both in-domain and cross-domain settings (TS-SD~\citep{shi2021self}, TS2Vec~\citep{yue2022ts2vec}, Mixing-Up~\citep{wickstrom2022mixing}, TF-C~\citep{zhang2022self}, TS-TCC~\citep{eldele2021time}, TST~\citep{zerveas2021transformer}, SimMTM~\citep{dong2023simmtm}) using SleepEEG dataset as the pre-training dataset, are reported in \citet{dong2023simmtm}, except for results of TS-SD which are reported in \citet{zhang2022self}.
The results of baseline methods for transfer learning in both self-supervised and semi-supervised settings (Supervised, TS-TCC~\citep{eldele2021time}, CA-TCC~\citep{eldele2022self}), using FD dataset as the pre-training dataset, are reported in \citet{eldele2022self}.

\begin{itemize}[leftmargin=22pt, itemsep=4pt]
    \item TS-SD~\citep{shi2021self}: TS-SD utilizes a triplet similarity discrimination task to train a model. The objective is to determine which of the two TS is more similar to a given TS, with DTW employed as a means to define the similarity.
    \item Mixing-Up~\citep{wickstrom2022mixing}: Mixing-up generates new time series by mixing two time series, and predicts the mixing weights.
    \item TF-C~\citep{zhang2022self}: TF-C generates both time-based and frequency-based representations of time series and proposes a novel time-frequency consistency architecture. 
    \item SimMTM~\citep{dong2023simmtm}: SimMTM adopts the masked modeling paradigm to time series domain, where the goal is to reconstruct the original time series from multiple masked series.
\end{itemize}

\paragraph{Anomaly detection: }
The results of all baseline methods for the anomaly detection task (SPOT~\citep{siffer2017anomaly}, DSPOT~\citep{siffer2017anomaly}, DONUT~\citep{xu2018unsupervised}, SR~\citep{ren2019time}, FFT~\citep{rasheed2009fourier}, Twitter-AD~\citep{vallis2014novel}, Luminol~\citep{luminol}, TS2Vec~\citep{yue2022ts2vec}) are reported in \citet{yue2022ts2vec}.
\begin{itemize}[leftmargin=22pt, itemsep=4pt]
    \item SPOT~\citep{siffer2017anomaly}: SPOT is a novel outlier detection approach for streaming univariate time series, based on Extreme Value Theory, which does not rely on pre-set thresholds, assumes no distribution, and only requires a single parameter to control the number of false positives.
    \item DONUT~\citep{xu2018unsupervised}: DONUT is an unsupervised anomaly detection algorithm based on variational autoencoder.
    \item SR~\citep{ren2019time}: SR is a time-series anomaly detection algorithm that is based on the Spectral Residual (SR) model and Convolutional Neural Network (CNN), where the SR model is borrowed from visual saliency detection and combined with CNN to improve its performance.
    \item FFT~\citep{rasheed2009fourier}: FFT uses fast fourier transform to detect the areas with high frequency change.
    \item Twitter-AD~\citep{vallis2014novel}: Twitter-AD automatically detects long-term anomalies in cloud data by identifying anomalies in application and system metrics.
    \item Luminol~\citep{luminol}: Luminol is a Python library for time series data analysis that provides two main functionalities - anomaly detection and correlation - and can be utilized to investigate the potential causes of anomalies.
\end{itemize}

\section{Implementation Details}
The table of hyperparameter settings that we utilized can be found in Table~\ref{tbl:hyperparameters}. We made use of five hyperparameters: $\tau_{I}$, $\tau_{T}$, $\lambda$, batch size (bs), and learning rate (lr). For semi-supervised classification and transfer learning, we set the weight decay to 3e-4, $\beta_1$ = 0.9, and $\beta_2$ = 0.
The number of optimization iterations for classification and anomaly detection tasks is set to 200 for datasets with a size less than 100,000; otherwise, it is set to 600. Additionally, the training epochs for semi-supervised classification are set to 80, while for transfer learning, it is set to 40.

Since we utilized soft contrastive loss as an auxiliary loss for TS-TCC and CA-TCC, which are the methods involved in solving semi-supervised classification and transfer learning tasks, we introduced an additional hyperparameter $\lambda_{aux}$ to control the contribution of the auxiliary loss to the final loss.

\begin{table*}[h]
\centering
\captionsetup{justification=centering}
\begin{adjustbox}{max width=0.95\textwidth}
\begin{NiceTabular}{c|c|c|c|c|c}
\toprule 
 & \multicolumn{2}{c}{Classification / Forecasting} & Semi-supervised classification & Transfer learning & Anomaly detection  \\
\midrule 
$\tau_{I}$ & \multicolumn{2}{c}{[1, 2, 3, 4, 5, 10, 20]} & \multicolumn{3}{c}{[10, 20, 30, 40, 50]}  \\
\midrule 
$\tau_{T}$ & \multicolumn{2}{c}{[0.5, 1.0, 1.5, 2.0, 2.5]} & \multicolumn{3}{c}{[1.5, 2.0, 2.5]}  \\
\midrule 
$\lambda$ & \multicolumn{2}{c}{0.5} & \multicolumn{2}{c}{[0.3, 0.5]} & 0.5 \\
\midrule 
$\lambda_{aux}$ & \multicolumn{2}{c}{-} & \multicolumn{2}{c}{[0.1, 0.3, 0.5]} & - \\
\midrule 
$\text{bs}$ & \multicolumn{2}{c}{8} & \multicolumn{2}{c}{16} & 4 (yahoo) / 8 (kpi) \\
\midrule 
$\text{lr}$ & \multicolumn{2}{c}{0.001} & \multicolumn{2}{c}{0.0003} & 0.001  \\
\bottomrule
\end{NiceTabular}
\end{adjustbox}
\centering
\caption{Hyperparameter settings for various tasks} 
\label{tbl:hyperparameters}
\end{table*}

\section{Probabilistic Interpretation of Soft Contrastive Losses} \label{supp_KL}
Inspired by the fact that the contrastive loss can be interpreted as the cross-entropy loss with virtual labels defined per batch, or equivalently, the KL divergence of the predicted softmax probability from the virtual label or hard assignment~\citep{lee2021mix}, we define a softmax probability of the relative similarity out of all similarities considered when computing the loss, and interpret our soft contrastive losses as a weighted sum of the cross-entropy losses.
In this section, we show that the proposed contrastive loss can also be seen as the scaled KL divergence of the predicted softmax probabilities from the normalized soft assignments, where the scale is the sum of soft assignments.
When hard assignment is applied, the loss becomes the standard contrastive loss, which is often called InfoNCE~\citep{oord2018representation}.

\subsection{Probabilistic Interpretation of Soft Instance-Wise Contrastive Loss}
To simplify indexing, we extend soft assignments to incorporate the positive sample and anchor itself:
\begin{align}
    w^{\prime}_{I}(i,i^{\prime})=
    \begin{cases}
        0, & \text{if } i=i^{\prime}; \\
        1, & \text{if } i\neq i^{\prime} \text{ and } i\equiv i^{\prime} (\text{mod } N); \\
        w_I(i,i^{\prime} \text{ mod } N), & \text{otherwise;}
    \end{cases}
\end{align}
and let $q_{I}(i,i^{\prime})= w^{\prime}_{I}(i,i^{\prime})/Z_I$ be its normalization, where $Z_I  = \sum_{j=1}^{2N} w^{\prime}_{I}(i,j)$ is the partition function.
Then, we can rewrite the proposed soft instance-wise contrastive loss as follows:
\begin{align*}
\ell_{I}^{(i, t)} &= - \log p_{I}((i,i+N),t) - \sum_{j=1,j \neq \{i, i+N\}}^{2N} w_{I}(i,j \text{ mod } N) \cdot \log p_{I}((i,j),t) \\
&= - \sum_{j=1}^{2N} w^{\prime}_{I}(i,j) \cdot \log p_{I}((i,j),t) \\
&= - Z_I  \cdot \sum_{j=1}^{2N} \frac{w^{\prime}_{I}(i,j)}{Z_I} \cdot \log p_{I}((i,j),t) \\
&= Z_I  \cdot \sum_{j=1}^{2N} q_{I}(i,j) \cdot \log \frac{q_{I}(i,j)}{p_{I}((i,j),t)} - \underset{\text{= constant}}{\underline{q_{I}(i,j) \log q_{I}(i,j)}}.
\stepcounter{equation}\tag{\theequation}\label{eqn:inst_loss_2}
\end{align*}
Let $Q_I$ and $P_I$ be the probability distributions of $q_{I}(i,j)$, and $p_{I}((i,j),t)$, respectively.
Then, we can rewrite the above loss as:
\begin{equation}
{
    \ell_{I}^{(i, t)} = Z_I  \cdot KL ( Q_{I} || P_I ) + \text{const},
}
\end{equation}
which is the scaled KL divergence of the predicted softmax probability from the soft assignments.

\subsection{Probabilistic Interpretation of Soft Temporal Contrastive Loss}
To simplify indexing, we extend soft assignments to incorporate the positive sample and anchor itself:
\begin{align}
    w^{\prime}_{T}(t,t^{\prime})=
    \begin{cases}
        0, & \text{if } t=t^{\prime}; \\
        1, & \text{if } t\neq t^{\prime} \text{ and } t\equiv t^{\prime} (\text{mod } T); \\
        w_T(t,t^{\prime} \text{ mod } T), & \text{otherwise;}
    \end{cases}
\end{align}
and let $q_{T}(t,t^{\prime})= w^{\prime}_{T}(t,t^{\prime})/Z_T$ be its normalization, where $Z_T  = \sum_{s=1}^{2T} w^{\prime}_T(t,s)$ is the partition function.
Then, we can rewrite the proposed soft temporal contrastive loss as follows:
\begin{align*}
\ell_{T}^{(i, t)} &= - \log p_{T}(i,(t,t+T)) - \sum_{s=1,s \neq \{t, t+T\}}^{2T} w_{T}(t,s \text{ mod } N) \cdot \log p_{T}(i,(t,s)) \\
&= - \sum_{s=1}^{2T} w^{\prime}_T(t,s) \cdot \log p_{T}(i,(t,s)) \\
&= - Z_T  \cdot \sum_{s=1}^{2T} \frac{w^{\prime}_T(t,s)}{Z_T} \cdot \log p_{T}(i,(t,s)) \\
&= Z_T  \cdot \sum_{s=1}^{2T} q_T(t,s) \cdot \log \frac{q_T(t,s)}{p_{T}(i,(t,s))} - \underset{\text{= constant}}{\underline{q_{T}(t,s) \log q_{T}(t,s)}}.
\stepcounter{equation}\tag{\theequation}\label{eqn:temp_loss_2}
\end{align*}
Let $Q_T$ and $P_T$ be the probability distributions of $q_{T}(t,s)$, and $p_{T}(i,(t,s))$, respectively.
Then, we can rewrite the above loss as:
\begin{equation}
{
    \ell_{T}^{(i, t)} = Z_T  \cdot KL ( Q_{T} || P_T ) + \text{const},
}
\end{equation}
which is the scaled KL divergence of the predicted softmax probability from the soft assignments.
These answer to a concern that targets are fixed while the predicted softmax probabilities are relative to the samples in the batch:
the formulation with fixed targets is proportional to the formulation with relative targets, and their difference is only in the optimization speed by the scale $Z_I$ and $Z_T$.

\newpage
\section{Hierarchical Soft Temporal Contrastive Loss}
\begin{wrapfigure}{r}{0.40\textwidth}
  \centering
     \begin{adjustbox}{max width=1.00\textwidth}
    \begin{NiceTabular}{c|c}
        \toprule
        Sharpness & Avg. Acc.(\%) \\
        \midrule
        $\tau_{T}$  & 83.3\\
        $m^{k}\cdot \tau_{T}$  & \textbf{83.7}\\
        \bottomrule
    \end{NiceTabular}    
    \end{adjustbox}
  \captionsetup{type=table}
  \caption{Effect of hierarchical $\tau_{T}$}
  \label{tbl:hierarchy_abl}
\end{wrapfigure}

In our approach, we adopt the hierarchical contrastive loss proposed in TS2Vec~\citep{yue2022ts2vec}, where we apply max-pooling on the representations along the temporal axis and contrastive learning is performed at each level.
However, as max pooling proceeds, semantic similarities between adjacent time step decrease, so the sharpness needs to be adjusted based on the hierarchy depth and kernel size.

To address this, we increase the value of sharpness in soft temporal contrastive loss as the depth of the network increases, thereby reflecting the hierarchy of the time series. 
That is, we use $m^{k}\cdot \tau_{T}$ instead of $\tau_{T}$ for the sharpness value in soft temporal contrastive loss, where $m$ is the kernel size of pooling layers and $k$ is the depth. 
For all datasets, we set $m$ to 2, while the value of $k$ depends on the length of each specific dataset.
We conducted an ablation study to assess the effect of using hierarchical sharpness, by comparing the performance of using hierarchical sharpness ($m^{k}\cdot \tau_T$) against a constant sharpness ($\tau_T$) using 128 datasets in UCR archive~\citep{dau2019ucr}. To solely observe the effect of hierarchical temporal contrastive loss, we employ the original hard instance-wise contrastive loss for this experiment.
The results presented in Table~\ref{tbl:hierarchy_abl} demonstrate that increasing $\tau_{T}$ as the depth of the network increases leads to improved performance.

\section{Design for Instance-Wise Contrastive Loss}
\begin{wrapfigure}{r}{0.40\textwidth}
  \centering
    \begin{adjustbox}{max width=\textwidth}
    \begin{NiceTabular}{c|c} \toprule
        Method & Avg. Acc.(\%) \\  
        \midrule 
        w/o kernel & 79.1 \\
        Gaussian & 82.6  \\
        Laplacian & 83.1 \\
        Sigmoid & \textbf{83.9} \\
        \bottomrule
    \end{NiceTabular}
    \end{adjustbox}
  \captionsetup{type=table}
  \caption{Design for instance-wise CL}
  \label{tbl:instfunction_abl}
\end{wrapfigure}
In this study, we explore different options for the soft assignments used in the soft instance-wise contrastive loss: without kernel (\textit{w/o kernel}), Laplacian kernel, and Gaussian kernel. 
For \textit{w/o kernel}, we use $w_I\left(i,j \right) = 1-D(x_i,x_j)$, where $D$ is a min-max normalized distance metric.
For the Laplacian kernel, we use $w_I\left(i,j \right) = \exp\left(-\frac{D(x_i,x_j)}{\sigma}\right)$, and
for the Gaussian kernel, we use $w_I\left(i,j \right)=\exp \left(-\frac{\left( D(x_i,x_j) \right)^2}{2 \sigma^2}\right)$, where $\sigma$ is a hyperparameter. 
For all kernels, sample pairs with a lower distance tends to have soft assignments closer to one. 
We conduct an ablation study by comparing the performance of using the above kernels to model the soft assignments using the UCR archive datasets, and the results are presented in Table~\ref{tbl:instfunction_abl}.
For this experiment, we employ the original hard temporal contrastive loss to solely observe the effect of the functions used for the instance-wise contrastive loss.

\section{Contrastive Learning for Anomaly Detection Task} \label{supp_AD}
Table~\ref{tbl:cl_ad} indicates that employing only temporal contrastive loss, while excluding instance-wise contrastive loss, yields better performance in the majority of hard CL and soft CL settings for anomaly detection tasks. This can be attributed to the nature of the anomaly detection task, which involves detecting anomalies within a time series, and is less concerned with other time series.

 \begin{table*}[h]
    \centering
    \captionsetup{justification=centering}
    \begin{subtable}[h]{0.48\textwidth}
        \centering
         \begin{adjustbox}{max width=1.00\textwidth}
          \begin{NiceTabular}{cc|ccc|ccc} 
          \toprule
          \multicolumn{2}{c}{\multirow{2}{*}{ }} & \multicolumn{3}{c}{\text{Yahoo}} & \multicolumn{3}{c}{\text{KPI}} \\
          \cmidrule(lr){3-5} \cmidrule(lr){6-8} 
           & & $\text{F}_1$ & \text{Prec.} & \text{Rec.} & $\text{F}_1$ & \text{Prec.} & \text{Rec.} \\
          \midrule 
          \multirow{2}{*}{TS2Vec} & \text{w/ inst} & \textbf{72.4} & \textbf{69.3} & 75.7 & 67.6 & 90.9 & 53.7 \\
          & \text{w/o inst} & 71.8 & 67.6 & \textbf{76.5} & \textbf{68.3} & \textbf{90.9} & \textbf{54.6} \\
          \midrule
          \multirow{2}{*}{+ Ours} & \text{w/ inst} & 71.2 & 67.8 & 74.9 & 66.4 & \textbf{94.3} & 51.4 \\
          & \text{w/o inst} & \textbf{74.2} & \textbf{72.2} & \textbf{76.5} & \textbf{70.1} & 91.6 & \textbf{57.0} \\
          \bottomrule
          \end{NiceTabular}
        \end{adjustbox}
        \caption{Results of AD task on normal setting}
        \label{tbl:cl_ad1}
    \end{subtable}
    \hspace{0.1cm}
    \begin{subtable}[h]{0.48\textwidth}
        \centering
         \begin{adjustbox}{max width=1.00\textwidth}
          \begin{NiceTabular}{cc|ccc|ccc} 
          \toprule
          \multicolumn{2}{c}{\multirow{2}{*}{ }} & \multicolumn{3}{c}{\text{Yahoo}} & \multicolumn{3}{c}{\text{KPI}} \\
          \cmidrule(lr){3-5} \cmidrule(lr){6-8} 
           & & $\text{F}_1$ & \text{Prec.} & \text{Rec.} & $\text{F}_1$ & \text{Prec.} & \text{Rec.} \\
          \midrule 
          \multirow{2}{*}{TS2Vec} & \text{w/ inst} & 74.0 & 70.7 & \textbf{77.6} & 68.9 & \textbf{89.3} & 56.2 \\
          & \text{w/o inst} & \textbf{75.5} & \textbf{73.6} & 77.4 & \textbf{69.7} & 88.8 & \textbf{57.4} \\
          \midrule
          \multirow{2}{*}{+ Ours} & \text{w/ inst} & 74.6 & 72.1 & \textbf{77.5} & 69.0 & \textbf{92.1} & 56.2 \\
          & \text{w/o inst} & \textbf{76.2} & \textbf{75.3} & 77.3 & \textbf{69.7} & \textbf{92.1} & \textbf{57.4} \\
          \bottomrule
          \end{NiceTabular}
        \end{adjustbox}
        \caption{Results of AD task on cold-start setting}
        \label{tbl:cl_ad2}
    \end{subtable}
    \setlength{\tabcolsep}{0.1pt}
    \caption{Results of anomaly detection task by the use of instance-wise contrastive loss.}
    \label{tbl:cl_ad}
\end{table*}

\newpage
\section{Time Series Forecasting}
The tasks mentioned in the main paper, except for anomaly detection, can be classified as high-level tasks, which requires capturing instance-wise representations.
High-level tasks generally perform better with CL methods than with masked modeling methods~\citep{dong2023simmtm, huang2022contrastive,xie2022simmim}. 
However, we can perform low-level tasks such as time series forecasting, when using encoder architectures that can obtain representations of each timestamp.

For TS forecasting, we apply SoftCLT to both TS2Vec and CoST \citep{woo2022cost}. 
Capturing temporal information within time series is crucial for time series forecasting, so we use soft CL in two ways for TS2Vec: by adopting only temporal contrastive loss and by using both temporal and instance-wise contrastive loss.
For the experiment, we use four datasets, ETTh1, ETTh2, ETTm1, and electricity dataset~\citep{zhou2021informer} for TS2Vec, and ETTh1, ETTh2, ETTm1, and Weather dataset~\citep{zhou2021informer} for CoST, under both univariate and multivariate settings.
Table~\ref{tbl:datastat_forecast} describes the summary of the statistical information for the five datasets.
As demonstrated in Table~\ref{tbl:tsforecasting} and Table~\ref{tbl:tsforecasting2}, our method results in performance gains compared to hard CL in both univariate and multivariate TS forecasting for both TS2Vec and CoST.

\begin{table*}[h]
\vspace{15pt}
\centering
\begin{adjustbox}{max width=0.88\textwidth}
    \begin{NiceTabular}{c|c|c|c}
    \toprule
    Datasets & Channels & Prediction Length & Samples \\
    \midrule 
    $\mathrm{ETTh}_1$, $\mathrm{ETTh}_2$ & 7 & \multirow{3}{*}{\{24,48,168,336,720\}} & 8640 / 2880 / 2880 \\
    Electricity & 321 &  & 15782 / 5261 / 5261 \\
    Weather & 21 &  & 36792 / 5271 / 10540 \\
    \midrule 
    $\textrm{ETTm}_1$ & 7 & \{24,48,96,288,672\} & 34560 / 11520 / 11520 \\
    \bottomrule
    \end{NiceTabular}
\end{adjustbox}
\caption{Four datasets used for time series forecasting, organized in the format of train/valid/test.}
\label{tbl:datastat_forecast}
\end{table*}

\begin{table*}[h]
\vspace{15pt}
\centering
\captionsetup{justification=centering}
\begin{adjustbox}{max width=0.999\textwidth}
    \begin{NiceTabular}{c|c|cccc|cccc|cccc|cccc} 
    \toprule
     &  & \multicolumn{8}{c}{Univariate forecasting} & \multicolumn{8}{c}{Multivariate forecasting} \\
     \cmidrule(l{0pt}r{0pt}){3-18}
     &  &  \multicolumn{4}{c}{w/ instance-wise CL} &  \multicolumn{4}{c}{w/o instance-wise CL} & \multicolumn{4}{c}{w/ instance-wise CL} & \multicolumn{4}{c}{w/o instance-wise CL} \\
     \cmidrule(l{0pt}r{0pt}){3-18}
     & & \multicolumn{2}{c}{TS2Vec} & \multicolumn{2}{c}{+ Ours} & \multicolumn{2}{c}{TS2Vec} & \multicolumn{2}{c}{+ Ours} & \multicolumn{2}{c}{\text {TS2Vec}} & \multicolumn{2}{c}{+ Ours} & \multicolumn{2}{c}{\text {TS2Vec}} & \multicolumn{2}{c}{+ Ours} \\
    \cmidrule(l{0pt}r{0pt}){3-18} 
    Dataset &  H & MSE & \text { MAE } & \text { MSE } & \text { MAE } & \text { MSE } & \text { MAE } & \text { MSE } & \text { MAE } & \text { MSE } & \text { MAE } & \text { MSE } & \text { MAE } & \text { MSE } & \text { MAE } & \text { MSE } & \text { MAE } \\
    \midrule 
    \multirow{6}{*}{$\mathrm{ETTh}_1$} & 24 & 0.042 & \textbf{0.152} & \textbf{0.041} & 0.156 & 0.046 & 0.164 & \textbf{0.045} & \textbf{0.161} & 0.568 & 0.513& \textbf{0.554} & \textbf{0.506} & 0.568 &0.525 & \textbf{0.554} &\textbf{0.510}\\
     & 48 & 0.067 & 0.197 & \textbf{0.064} & \textbf{0.194} & \textbf{0.079} & \textbf{0.216} & 0.080 & 0.218 & 0.607 & 0.538 & \textbf{0.595} & \textbf{0.532} &0.617 &0.557 &\textbf{0.593}&\textbf{0.542}\\
    & 168 & 0.154 & 0.304 & \textbf{0.144} & \textbf{0.293} & 0.153 & 0.302 & \textbf{0.144} & \textbf{0.291} & 0.742 & \textbf{0.622} & \textbf{0.737} & 0.624 &0.796 &0.664&\textbf{0.765}&\textbf{0.647}\\
     & 336 & 0.174 & 0.332 & \textbf{0.162} & \textbf{0.318} & 0.172 & 0.328 & \textbf{0.160} & \textbf{0.314} & 0.937 & 0.726 & \textbf{0.890} & \textbf{0.712} &1.024&0.777&\textbf{0.867}&\textbf{0.702}\\
     & 720 & 0.209 & 0.376 & \textbf{0.179} & \textbf{0.345} & 0.192 & 0.357 & \textbf{0.178} & \textbf{0.341} & 1.068 & 0.800 & \textbf{1.056} & \textbf{0.798} &1.063&0.801&\textbf{1.046}&\textbf{0.795}\\
     \cmidrule(l{0pt}r{0pt}){2-18}
   & Avg. & 0.129 & 0.272 & \textbf{0.120} & \textbf{0.261} & 0.128 & 0.273 & \textbf{0.121} & \textbf{0.265} & 0.784 & 0.640 & \textbf{0.766} & \textbf{0.634} &0.814&0.665&\textbf{0.765}&\textbf{0.639}\\
    \midrule 
    \multirow{6}{*}{$\mathrm{ETTh}_2$} & 24 & 0.090 & 0.230 & \textbf{0.086} & \textbf{0.224} & 0.090 & 0.229 & \textbf{0.088} & \textbf{0.226} & 0.373 & 0.465 & \textbf{0.370} & \textbf{0.462} & 0.371 & 0.462&\textbf{0.362}&\textbf{0.452}\\
    & 48 & 0.126 & 0.273 & \textbf{0.121} & \textbf{0.268} & 0.121 & 0.268 & \textbf{0.119} & \textbf{0.265} & 0.561 & 0.579 & \textbf{0.557} & \textbf{0.577} &0.548&0.571&\textbf{0.535}&\textbf{0.559}\\
    & 168 & 0.208 & 0.359 & \textbf{0.202} & \textbf{0.354} & 0.196 & 0.349 & \textbf{0.194} & \textbf{0.347} & \textbf{1.713} & \textbf{1.015} & \textbf{1.713} & 1.016 &1.693&1.024&\textbf{1.606}&\textbf{1.001}\\
    & 336 & 0.219 & 0.374 & \textbf{0.206} & \textbf{0.363} & 0.207 & 0.364 & \textbf{0.205} & \textbf{0.362} & 2.153 & 1.167 & \textbf{2.061} & \textbf{1.147} &2.096&1.172&\textbf{1.973}&\textbf{1.135}\\
   & 720 & 0.221 & 0.381 & \textbf{0.216} & \textbf{0.377} & 0.217 & 0.377 & \textbf{0.215} & \textbf{0.376} & 2.437 & 1.299 & \textbf{2.394} & \textbf{1.275} &2.464&1.319&\textbf{2.297}&\textbf{1.259}\\
   \cmidrule(l{0pt}r{0pt}){2-18}
   & Avg. & 0.173 & 0.323 & \textbf{0.166} & \textbf{0.316} & 0.166 & 0.317 & \textbf{0.164} & \textbf{0.315} & 1.447 & 0.905 & \textbf{1.441} & \textbf{0.895} &1.434 &0.910&\textbf{1.355}&\textbf{0.881}\\
    \midrule 
    \multirow{6}{*}{$\mathrm{ETTm}_1$} & 24 & 0.016 & 0.093 & \textbf{0.014} & \textbf{0.088} & 0.015 & 0.092 & \textbf{0.014} & \textbf{0.090} & 0.459 & 0.449 & \textbf{0.418} & \textbf{0.426} &0.428& 0.430&\textbf{0.421}&\textbf{0.423}\\
    & 48 & 0.029 & 0.128 & \textbf{0.027} & \textbf{0.124} & 0.028 & 0.126 & \textbf{0.027} & \textbf{0.124} & 0.608 & 0.521 & \textbf{0.567} & \textbf{0.501} &0.587&0.512&\textbf{0.568}&\textbf{0.501}\\
    & 96 & 0.044 & 0.158 & \textbf{0.041} & \textbf{0.155} & \textbf{0.048} & \textbf{0.166} & \textbf{0.048} & \textbf{0.166} & 0.597 & 0.532 & \textbf{0.591} & \textbf{0.530} &0.623&0.544&\textbf{0.595}&\textbf{0.524}\\
    & 288 & 0.103 & 0.246 & \textbf{0.093} & \textbf{0.232} & \textbf{0.113} & \textbf{0.258} & 0.115 & 0.260 & 0.670 & 0.586& \textbf{0.647} & \textbf{0.577} &0.704&0.600&\textbf{0.659}&\textbf{0.580}\\
    & 672 & 0.155 & 0.298 & \textbf{0.135} & \textbf{0.283} & 0.163 & 0.313 & \textbf{0.160} & \textbf{0.311} & 0.750 & 0.639 & \textbf{0.743} & \textbf{0.637} &0.797&0.659&\textbf{0.753}&\textbf{0.642}\\
    \cmidrule(l{0pt}r{0pt}){2-18}
   & Avg. & 0.069 & 0.185 & \textbf{0.062} & \textbf{0.176} & \textbf{0.073} & 0.191 & \textbf{0.073} & \textbf{0.190} & 0.617 & 0.545 & \textbf{0.593} & \textbf{0.534} &0.628&0.549&\textbf{0.599}&\textbf{0.534}\\
    \midrule 
    \multirow{6}{*}{\text { Electricity }} & 24 & 0.259 & 0.291 & \textbf{0.251} & \textbf{0.284} & 0.268 & 0.299 & \textbf{0.252} & \textbf{0.286} & \textbf{0.285} & 0.375 & 0.286 &\textbf{0.375} &0.317& 0.400 &\textbf{0.315}&\textbf{0.398}\\
    & 48 & 0.309 & 0.323 & \textbf{0.304} & \textbf{0.317} & 0.326 & 0.350 & \textbf{0.306} & \textbf{0.323} & \textbf{0.308} & 0.391 & \textbf{0.308} &\textbf{0.391}&0.340& 0.415 &\textbf{0.338}&\textbf{0.413}\\
    & 168 & 0.426 & 0.397 & \textbf{0.418} & \textbf{0.391} & 0.446 & 0.431 & \textbf{0.425} & \textbf{0.401} & 0.335 & 0.411 & \textbf{0.334} &\textbf{0.411}&0.364& 0.432 &\textbf{0.362}&\textbf{0.430}\\
    & 336 & 0.567 & 0.484 & \textbf{0.560} & \textbf{0.479} & 0.589 & 0.524 & \textbf{0.571} & \textbf{0.494} & 0.352 & 0.424 & \textbf{0.351} &\textbf{0.424}&0.380& 0.443 &\textbf{0.377}&\textbf{0.441}\\
    & 720 & 0.860 & 0.650 & \textbf{0.858} & \textbf{0.645} & 0.882 & 0.700 & \textbf{0.879} & \textbf{0.685} & \textbf{0.378} & 0.442 & \textbf{0.378} &\textbf{0.442}&0.403& 0.459 &\textbf{0.401}&\textbf{0.457}\\
    \cmidrule(l{0pt}r{0pt}){2-18}
   & Avg. & 0.484 & 0.429 & \textbf{0.478} & \textbf{0.423} & 0.502 & 0.461 & \textbf{0.451} & \textbf{0.438}  & 0.332 & \textbf{0.409} & \textbf{0.331} & \textbf{0.409} &0.361 & 0.430 &\textbf{0.359}&\textbf{0.428}\\
    \bottomrule
    \end{NiceTabular}
\end{adjustbox}
\caption{Results of univariate and multivariate time series forecasting.}
\label{tbl:tsforecasting}
\end{table*}   

\begin{table*}[h]
\centering
\captionsetup{justification=centering}
\begin{adjustbox}{max width=0.9\textwidth}
    \begin{NiceTabular}{c|c|cccc|cccc} 
    \toprule
     &  &  \multicolumn{4}{c}{Univariate forecasting} &  \multicolumn{4}{c}{Multivariate forecasting} \\
     \cmidrule(l{0pt}r{0pt}){3-10}
     & & \multicolumn{2}{c}{CoST} & \multicolumn{2}{c}{+ Ours} & \multicolumn{2}{c}{CoST} & \multicolumn{2}{c}{+ Ours} \\
    \cmidrule(l{0pt}r{0pt}){3-10} 
    Dataset &  H & MSE & \text { MAE } & \text { MSE } & \text { MAE } & \text { MSE } & \text { MAE } & \text { MSE } & \text { MAE } \\
    \midrule 
    \multirow{6}{*}{$\mathrm{ETTh}_1$} 
    & 24 & \textbf{0.040} & \textbf{0.152} & \textbf{0.040} & \textbf{0.152} & 0.386 & 0.429 & \textbf{ 0.377 } & \textbf{ 0.422} \\
     & 48 & 0.064 & \textbf{ 0.186 } & \textbf{ 0.061 } & \textbf{ 0.186 } & 0.437 & 0.464 & \textbf{ 0.428 } & \textbf{ 0.455} \\
    & 168 & 0.097 & 0.236 & \textbf{ 0.093 } & \textbf{ 0.230 } & 0.643 & 0.582 & \textbf{ 0.626 } & \textbf{ 0.571} \\
     & 336 & 0.112 & 0.258 & \textbf{ 0.110 } & \textbf{ 0.254 } & 0.812 & 0.679 & \textbf{ 0.773 } & \textbf{ 0.660} \\
     & 720 & 0.158 & 0.316 & \textbf{ 0.155 } & \textbf{ 0.314 } & 0.970 & 0.771 & \textbf{ 0.891 } & \textbf{0.744} \\
     \cmidrule(l{0pt}r{0pt}){2-10}
    & Avg. & 0.094 & 0.230 & \textbf{ 0.091 } & \textbf{ 0.227 } & 0.650 & 0.585 & \textbf{ 0.619 } & \textbf{ 0.570} \\
    \midrule 
    \multirow{6}{*}{$\mathrm{ETTh}_2$} 
    & 24 & 0.079 & 0.207 & \textbf{ 0.078 } & \textbf{ 0.206 } & 0.480 & 0.525 & \textbf{ 0.472 } & \textbf{ 0.522} \\
    & 48 & 0.118 & 0.259 & \textbf{ 0.117 } & \textbf{ 0.257 } & 0.751 & 0.669 & \textbf{ 0.749 } & \textbf{ 0.667} \\
    & 168 & 0.189 & 0.339 & \textbf{ 0.179 } & \textbf{ 0.332 } & 1.613 & 1.017 & \textbf{ 1.602 } & \textbf{ 1.009} \\
    & 336 & 0.206 & 0.360 & \textbf{ 0.201  } & \textbf{ 0.357 } & 1.807 & 1.078 & \textbf{ 1.800 } & \textbf{ 1.075} \\
   & 720 & 0.214 & 0.371 & \textbf{ 0.208 } & \textbf{ 0.369 } & 1.959 & 1.099 & \textbf{ 1.951 } & \textbf{ 1.090} \\
   \cmidrule(l{0pt}r{0pt}){2-10}
   & Avg. & 0.161 & 0.307 & \textbf{ 0.156 } & \textbf{ 0.304 } & 1.322 & 0.876 & \textbf{ 1.315 } & \textbf{ 0.872} \\
    \midrule 
    \multirow{6}{*}{$\mathrm{ETTm}_1$} 
    & 24 & 0.015 & 0.088 & \textbf{ 0.014 } & \textbf{ 0.087 } & 0.246 & 0.329 & \textbf{ 0.243 } & \textbf{ 0.326} \\
    & 48 & 0.025 & 0.117 & \textbf{ 0.024 } & \textbf{ 0.116 } & \textbf{ 0.331 } & \textbf{ 0.386 } & 0.332 & \textbf{ 0.386} \\
    & 96 & 0.038 & 0.147 & \textbf{ 0.036 } & \textbf{ 0.145 } & 0.378 & 0.419 & \textbf{ 0.376 } & \textbf{ 0.418} \\
    & 288 & 0.077 & 0.209 & \textbf{ 0.075 } & \textbf{ 0.204 } & \textbf{ 0.472 } &  0.486  & 0.474 & \textbf{ 0.484} \\
    & 672 & 0.113 & 0.257 & \textbf{ 0.105 } & \textbf{ 0.245 } & 0.620 & 0.574 & \textbf{ 0.613 } & \textbf{ 0.567} \\
    \cmidrule(l{0pt}r{0pt}){2-10}
   & Avg. & 0.054 & 0.164 & \textbf{ 0.051 } & \textbf{ 0.159 } & 0.409 & 0.439 & \textbf{ 0.407 } & \textbf{ 0.436} \\ 
    \midrule 
    \multirow{6}{*}{\text { Weather }}  
    & 24 & 0.096 & 0.213 & \textbf{ 0.095 } & \textbf{ 0.212 } & 0.299 & 0.361 & \textbf{ 0.298 } & \textbf{ 0.360} \\
    & 48 & 0.139 & 0.263 & \textbf{ 0.132 } & \textbf{ 0.260 } & 0.359 & 0.411 & \textbf{ 0.357 } & \textbf{ 0.410} \\
    & 168 & 0.213 & 0.338 & \textbf{ 0.208 } & \textbf{ 0.335 } &0.464 & \textbf{ 0.491 } & \textbf{ 0.463 } & \textbf{0.491} \\
    & 336 & 0.235 & 0.360 & \textbf{ 0.208 } & \textbf{ 0.357 } & 0.499 & 0.517 & \textbf{ 0.496 } & \textbf{ 0.513} \\
    & 720 & 0.251 & 0.375 & \textbf{ 0.231 } & \textbf{ 0.357 } & \textbf{ 0.535 } & \textbf{ 0.544 } & 0.538 & \textbf{ 0.544} \\
    \cmidrule(l{0pt}r{0pt}){2-10}
   & Avg. & 0.189 & 0.310 & \textbf{ 0.181 } & \textbf{ 0.307 } & 0.432 & 0.465 & \textbf{ 0.430 } & \textbf{ 0.463} \\
    \bottomrule
    \end{NiceTabular}
\end{adjustbox}
\caption{Results of univariate and multivariate time series forecasting for CoST.}
\label{tbl:tsforecasting2}
\vspace{10pt}
\end{table*}   

\newpage
\section{Instance-Wise Visualizations}
\begin{figure}[t]
    \centering
    \begin{minipage}{0.425\textwidth}
        \centering
        \includegraphics[width=\textwidth]{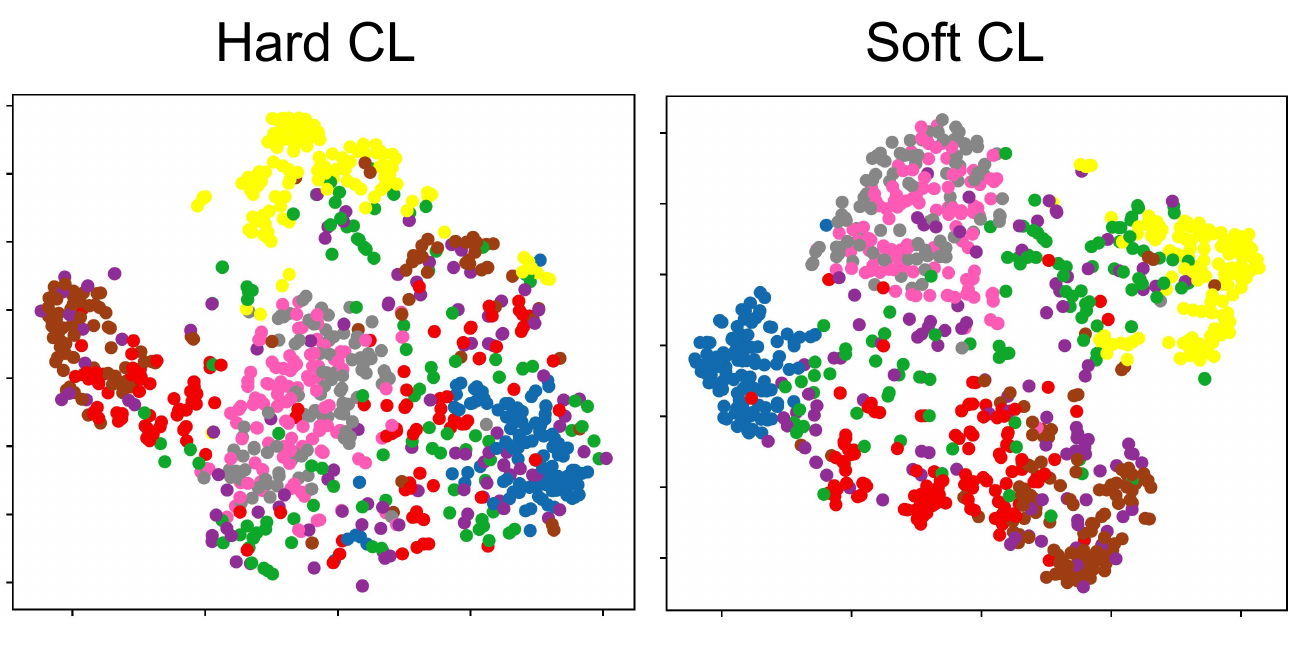} 
        \caption{Hard CL vs. Soft CL}
        \label{fig:clustering_temporal_appendix}
    \end{minipage} 
    \hfill
    \begin{minipage}{0.485\textwidth}
        \centering
        \includegraphics[width=\textwidth]{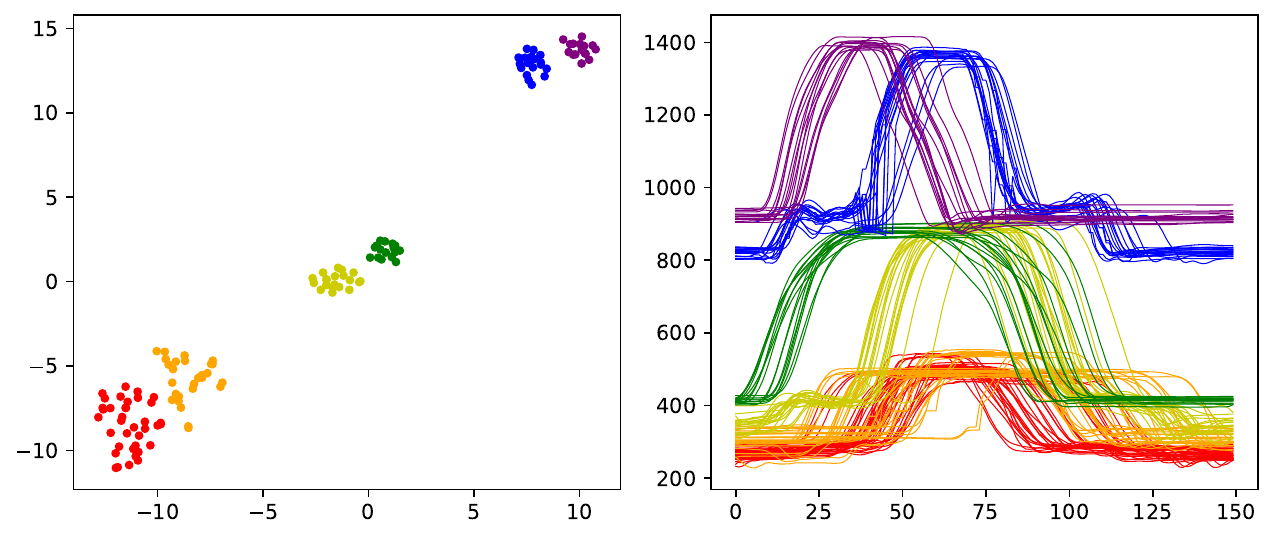} 
        \caption{Instance-wise visualizations }
        \label{fig:clustering_instance-wise_appendix}
    \end{minipage}
\end{figure}

\textbf{Hard CL vs. Soft CL. }
To assess the quality of instance-wise relationships captured by SoftCLT, we apply t-SNE~\citep{van2008visualizing} to visualize the instance-wise representations, which are representations of whole time series obtained by max-pooling the representations of all time stamps, to both hard and soft CL.
For this experiment, we apply our method to TS2Vec~\citep{yue2022ts2vec} with the UWaveGestureLibraryZ
dataset from UCR archive~\citep{dau2019ucr}.
The results shown in Figure~\ref{fig:clustering_temporal_appendix} demonstrate that soft CL finds more fine-grained neighborhood relationships and distinguishes them better than hard CL.

\textbf{Embedding space vs. Input space. }
To assess the relationship between the shape of time series and their positions in the embedding dimension, we employ t-SNE~\citep{van2008visualizing} to embed instance-wise representations of time series using the InsectEPGRegularTrain dataset from UCR archive~\citep{dau2019ucr}.
Figure~\ref{fig:clustering_instance-wise_appendix} illustrates the results, with the left panel displaying the points in the embedding space and the right panel presenting line plots of the original TS. The colors of the points and lines are assigned based on the distances with their neighbors in the embedding space.
From this figure, we observe that TS with the same color not only exhibit similar shapes, but also as the points in the embedding space move towards the upper right, the line plots of the original TS shift towards the upper left.
This demonstrates that our method effectively captures detailed neighborhood relationships while maintaining alignment between the distances in the embedding space and the original input space.

\newpage
\section{Transfer Learning Under Semi-supervised Settings}
\begin{wrapfigure}{r}{0.30\textwidth}
  \centering
  \includegraphics[width=0.30\textwidth]{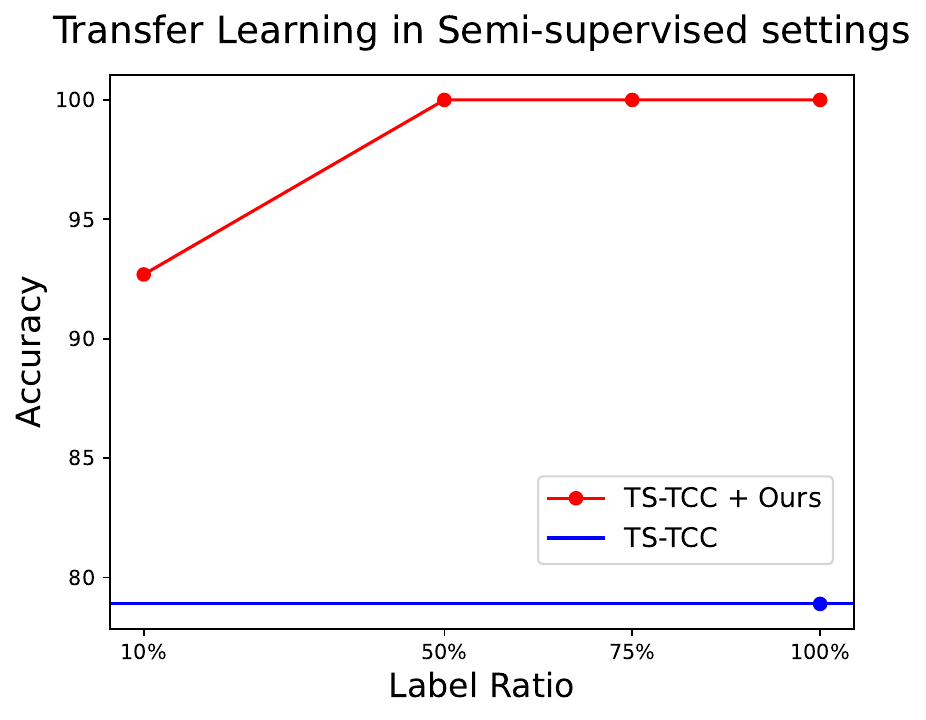} 
  \caption{TL results}
  \label{fig:TL_SemiSL}
\end{wrapfigure}
In this study, we perform transfer learning in the semi-supervised settings using SleepEEG~\citep{kemp2000analysis} and EMG~\citep{goldberger2000physiobank} datasets as the source and target datasets, respectively. Specifically, we apply our SoftCLT to TS-TCC under semi-supervised settings where we perform fine-tuning using partially labeled datasets.
Figure~\ref{fig:TL_SemiSL} presents the results, which indicate that by using only 10\% of labeled data with the soft CL framework (red line), we are able to achieve an accuracy of 92.69\%, which is approximately 15\% higher than the accuracy obtained from the hard CL framework (blue line) under fully supervised settings. Furthermore, using only 50\% of the labeled dataset allowed us to achieve 100\% accuracy, whereas the state-of-the-art performance using fully labeled datasets is 97.56\%.

\section{Effect of Distance Metrics by Time Series with Varying Length}
We compare the average accuracy of 128 UCR datasets, where 11 datasets have varying time-length, and the other 117 datasets have the same time-length. 
As shown in Table \ref{tbl:vary_nonvary_length}, DTW and TAM, both capable of comparing time series of different lengths using time warping, demonstrate better performance.

\begin{table*}[h]
\centering
  \centering
    \begin{adjustbox}{max width=0.84\textwidth}
    \begin{NiceTabular}{c|ccc|ccc}
        \toprule
        &\multicolumn{6}{c}{UCR datasets (Avg. Acc.(\%))} \\
        \midrule
        Temporal CL&\multicolumn{3}{c}{Hard}&\multicolumn{3}{c}{Soft} \\
        \midrule
      \multirow{2}{*}{TS length} & Non-Varying  & Varying  & Total  & Non-Varying & Varying & Total \\
 & (117/128) & (11/128) & (128/128) & (117/128) & (11/128) & (128/128) \\
        \midrule
        COS & 84.8 & 72.6  & 83.7 & 85.7 & 75.0  & 84.7  \\
        EUC & 85.1 & 73.3  & 83.9 & 85.8 & 73.9  & 84.8  \\
        DTW & 84.8 & 73.6  & 83.9 & 85.9 & 75.2  & 85.0  \\
        TAM & 85.0 & 73.4  & 83.9 & 85.9 & 75.3  & 85.0  \\
        \bottomrule
    \end{NiceTabular}
    \end{adjustbox}
  \caption{Effect of DTW on time series with varying/non-varying length.}
 \label{tbl:vary_nonvary_length}
\end{table*}

\section{Design Choices for Soft Temporal Contrastive Learning}
\begin{wrapfigure}{r}{0.44\textwidth}
  \centering
  \includegraphics[width=0.44\textwidth]{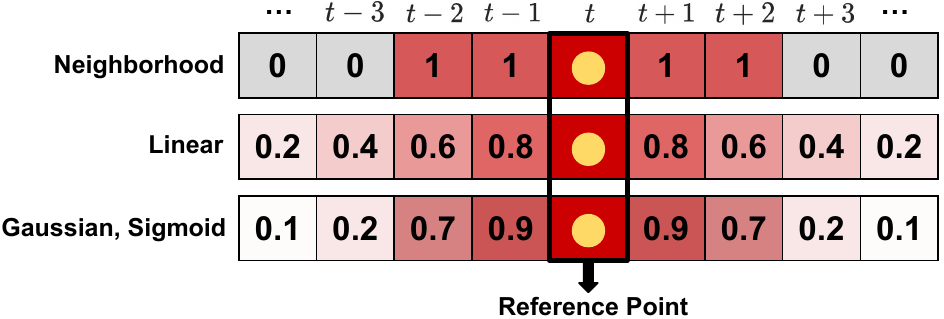} 
  \caption{Design for soft temporal CL}
  \label{fig:temporal_CL_designs}
  \vspace{-10pt}
\end{wrapfigure}
Various design choices can be considered for assigning soft labels in soft temporal contrastive learning. 
In this paper, we explore four different choices for the experiment, all of which assign high values to adjacent timestamps. 
Figure \ref{fig:temporal_CL_designs} illustrates these four different choices.
For \textbf{Neighbor}, \textbf{Gaussian}, and \textbf{Sigmoid}, 
we conduct a search for the optimal hyperparameter within the following range:
\begin{itemize}
\item \textbf{Neighbor}:  A certain range within the reference point, with 10\%, 30\%, 50\% of the sequence length.
\item \textbf{Gaussian}: Standard deviation values of [0.5, 1.0, 1.5, 2.0, 2.5].
\item \textbf{Sigmoid}: $\tau_T$ of [0.5, 1.0, 1.5, 2.0, 2.5].
\end{itemize}

\newpage
\section{Soft Contrastive Learning with Non-stationary Time Series} \label{supp_nonstationarity} 
As our proposed soft instance-wise CL method generates labels by considering the distances between the \textit{original} TS, global information from the entire TS is encapsulated within the representation of a single time step of that TS, which might enables to take account of non-stationarity, such as seasonality or distribution shifts present in the TS.

\begin{figure*}[h]
\centering
\includegraphics[width=0.8\textwidth]{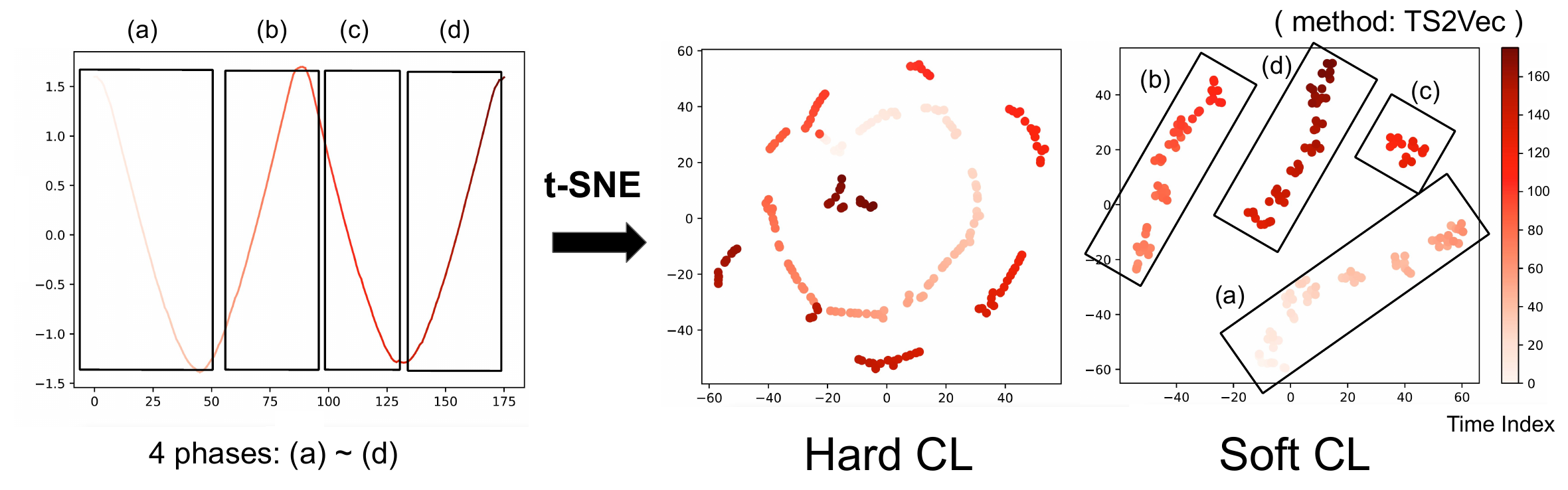} 
\caption{Temporal visualization of TS with seasonality.}
\label{fig:season}
\vspace{12pt}
\end{figure*}

\textbf{Time series with seasonality. }
Figure~\ref{fig:season} displays a single TS of Adiac data from the UCR archive~\citep{dau2019ucr} and its visualization of temporal representations obtained from TS2Vec (Hard CL) and TS2Vec applied with our method (Soft CL). 
Note that obvious seasonal pattern are observed from the left panel of the figure.
Each point in the right panel represents a representation of a single timestamp. 
The figure indicates that while hard CL fails to capture the seasonal patterns in the TS (as similar values, regardless of which phase they are in, are located closely), our proposed soft CL can grasp the global pattern, enabling it to capture the seasonal patterns (as similar values but with different seasonal phases are located differently).

\begin{figure*}[h]
\centering
\includegraphics[width=0.8\textwidth]{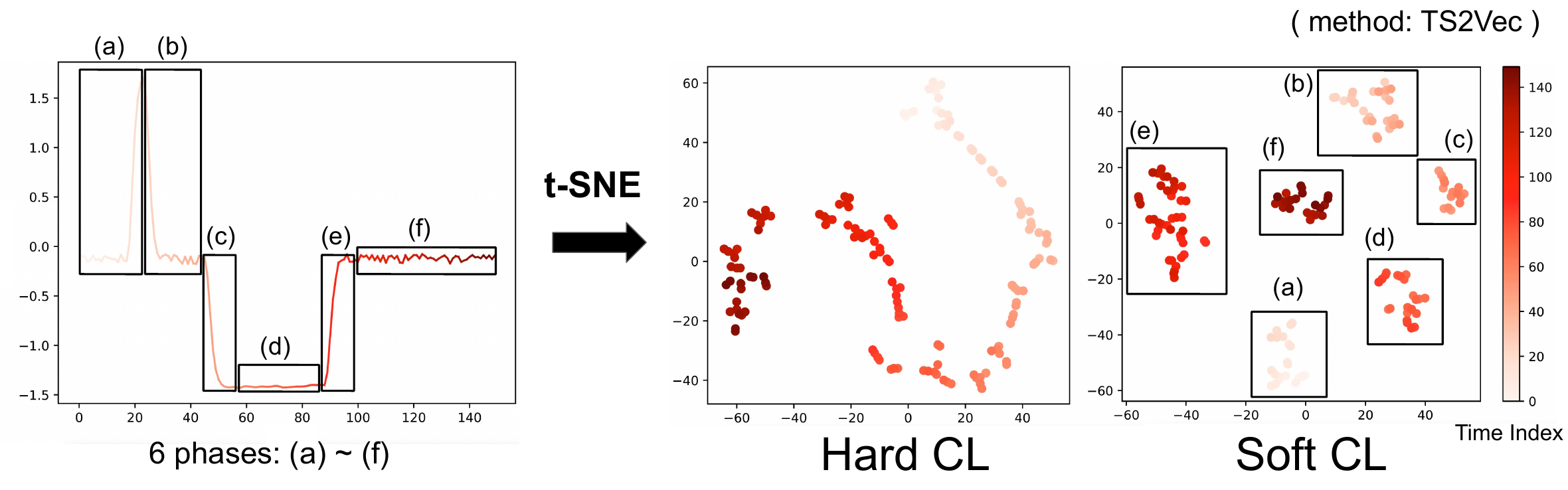} 
\caption{Temporal visualization of TS with distribution shift.}
\label{fig:distn_shift}
\end{figure*}

\textbf{Time series with distribution shift. }
Figure~\ref{fig:distn_shift} displays a single TS of EMD data from the UCR archive~\citep{dau2019ucr} and its visualization of temporal representations obtained from TS2Vec (Hard CL) and TS2Vec applied with our method (Soft CL). 
Note that distribution shifts exist in this data, and six different phases are observed from the left panel of the figure.
Each point in the right panel represents a representation of a single timestamp. The figure indicates that while hard CL fails to capture the sudden change in the TS (as all points are located gradually regardless of the sudden change in the TS), our proposed soft CL can detect such distribution shifts (as points before/after a certain change are clustered in groups).

\newpage
\section{Applying Soft Contrastive Learning to TNC}
In this section, we apply our SoftCLT to temporal neighborhood coding (TNC)~\citep{tonekaboni2021unsupervised}, which employs temporal CL by leveraging the local smoothness of a signal's generative process to define neighborhoods, or positive pairs, in time with stationary properties.
We apply our method to TNC by assigning soft temporal assignments based on the difference between the centroids of two time windows.
Note that TNC does not include instance-wise CL, so we only apply soft temporal CL.
To evaluate the effectiveness of this application, we conduct experiments using two datasets:
the simulation dataset constructed in TNC and HAR~\footnote{\url{https://archive.ics.uci.edu/dataset/240/human+activity+recognition+using+smartphones}}. 
The results are shown in Table~\ref{tab:TNC}, demonstrating that SoftCLT applied to TNC improves performance on both datasets in terms of accuracy (Acc.) and AUPRC.

\begin{table*}[h]
\centering
  \centering
    \vspace{10pt}
    \begin{adjustbox}{max width=1.0\textwidth}
    \begin{NiceTabular}{c|cc|cc}
        \toprule
        &\multicolumn{2}{c}{Simulation} & \multicolumn{2}{c}{HAR}\\
        \cmidrule{2-5}
        & Acc.(\%) & AUPRC & Acc.(\%) & AUPRC \\
        \midrule
        T-Loss & 76.66 & 0.78  & 63.60 & 0.71  \\
        CPC & 70.26 & 0.69  & 86.43 & 0.93  \\
        TNC & 97.52 & 0.996  & 88.21 & 0.940 \\
        \midrule
        TNC + Ours & \textbf{97.95} & \textbf{0.998} & \textbf{89.14} & \textbf{0.952} \\ 
        \bottomrule
    \end{NiceTabular}
    \end{adjustbox}
  \caption{Effect of soft temporal CL on TNC.}
 \label{tab:TNC}
\end{table*}